\documentclass[lettersize,journal]{IEEEtran}
\usepackage{amsmath,amsfonts}
\usepackage{algorithmic}
\usepackage{algorithm}
\usepackage{array}
\usepackage[caption=false,font=normalsize,labelfont=sf,textfont=sf]{subfig}
\usepackage{textcomp}
\usepackage{stfloats}
\usepackage{url}
\usepackage{verbatim}
\usepackage{graphicx}
\usepackage{cite}
\usepackage{tabularray}
\usepackage{booktabs} % 专业表格线
\usepackage{siunitx}  % 数字对齐
\usepackage{orcidlink} %作者后面加orcid号
\begin{document}

\title{MFGDiffusion: Mask-Guided Smoke Synthesis for Enhanced Forest Fire Detection}

\author{Guanghao Wu $^{\orcidlink{0009-0005-0393-6247}}$, Yunqing Shang $^{\orcidlink{ 0009-0006-5071-8725}}$, Chen Xu $^{\orcidlink{0000-0003-4163-6024}}$, Hai Song $^{\orcidlink{0009-0006-3201-306X}}$, Chong Wang $^{\orcidlink{0009-0008-5190-7792}}$, and Qixing Zhang$^{\orcidlink{0000-0002-8784-8674}}$

\thanks{ This work was financially supported by the National Natural Science Foundation of China under Grant No. 32471866, the Anhui Provincial Science and Technology Major Project under Grant No. 202203a07020017, the Central Government-Guided Local Science and Technology Development Funds of Anhui Province under Grant No. 2023CSJGG1100 and the Key Project of Emergency Management Department of China under Grant No. 2024EMST010101. This research was also supported by the advanced computing resources provided by the Supercomputing Center of the USTC.The authors gratefully acknowledge all of these supports. (Guanghao Wu and Yunqing Shang are co-first authors.)(Corresponding authors: Qixing Zhang.).}
% introduce author`s 
\thanks{ Guanghao Wu is with the School of Artificial Intelligence and Data Science, University of Science and Technology of China, Hefei, 230026, China (e-mail: wgh13783456627@mail.ustc.edu.cn).}
\thanks{ Yunqing Shang and Hai Song are with the Institute of Advanced Technology, University of Science and Technology of China, Hefei, 230026, China (e-mail: shang\_yq@mail.ustc.edu.cn; shlyyy@mail.ustc.edu.cn).}
\thanks{ Chen Xu, Chong Wang and Qixing Zhang are with the State Key Laboratory of Fire Science, University of Science and Technology of China, Hefei, 230026, China (e-mail: xuchenustc@mail.ustc.edu.cn;  wchong@ustc.edu.cn; qixing@ustc.edu.cn).}

}





\maketitle

\begin{abstract}
		Smoke is the first visible indicator of a wildfire.With the advancement of deep learning, image-based smoke detection has become a crucial method for detecting and preventing forest fires. However, the scarcity of smoke image data from forest fires is one of the significant factors hindering the detection of forest fire smoke. Image generation models offer a promising solution for synthesizing realistic smoke images. However, current inpainting models exhibit limitations in generating high-quality smoke representations, particularly manifesting as inconsistencies between synthesized smoke and background contexts. To solve these problems, we proposed a comprehensive framework for generating forest fire smoke images. Firstly, we employed the pre-trained segmentation model and the multimodal model to obtain smoke masks and image captions.Then, to address the insufficient utilization of masks and masked images by inpainting models, we introduced a network architecture guided by mask and masked image features. We also proposed a new loss function, the mask random difference loss, which enhances the consistency of the generated effects around the mask by randomly expanding and eroding the mask edges.Finally, to generate a smoke image dataset using random masks for subsequent detection tasks, we incorporated smoke characteristics and use a multimodal large language model as a filtering tool to select diverse and reasonable smoke images, thereby improving the quality of the synthetic dataset. Experiments showed that our generated smoke images are realistic and diverse, and effectively enhance the performance of forest fire smoke detection models. Code is available at \url{https://github.com/wghr123/MFGDiffusion}.
\end{abstract}


\section{Introduction}
\IEEEPARstart{D}{ue} to global warming, forest fires are becoming increasingly frequent, causing severe environmental damage and posing threats to human life and property safety. To mitigate the risk of forest fires, forest fire prevention efforts must not only focus on proactive prevention and management but also emphasize the rapid detection, early warning, and timely response to fires. In forest scenarios, smoke is a critical early indicator of fire. Smoke detection based on deep learning \cite{li20183d}, \cite{tao2019smoke}, \cite{cao2021effnet}, \cite{jing2023smokepose}, \cite{wang2024dpmnet} has emerged as a significant approach for early forest fire prevention. However, deep learning-based smoke detection models require a large and diverse dataset of smoke images. Currently, publicly available datasets for forest fire smoke are limited in quantity and quality, making it challenging to train highly accurate and generalizable smoke detection models.

    \par
\begin{figure*}[h] % htbp详见说明书，记得删除括号内容
		\centering % 居中
		\includegraphics[width=0.8\linewidth]{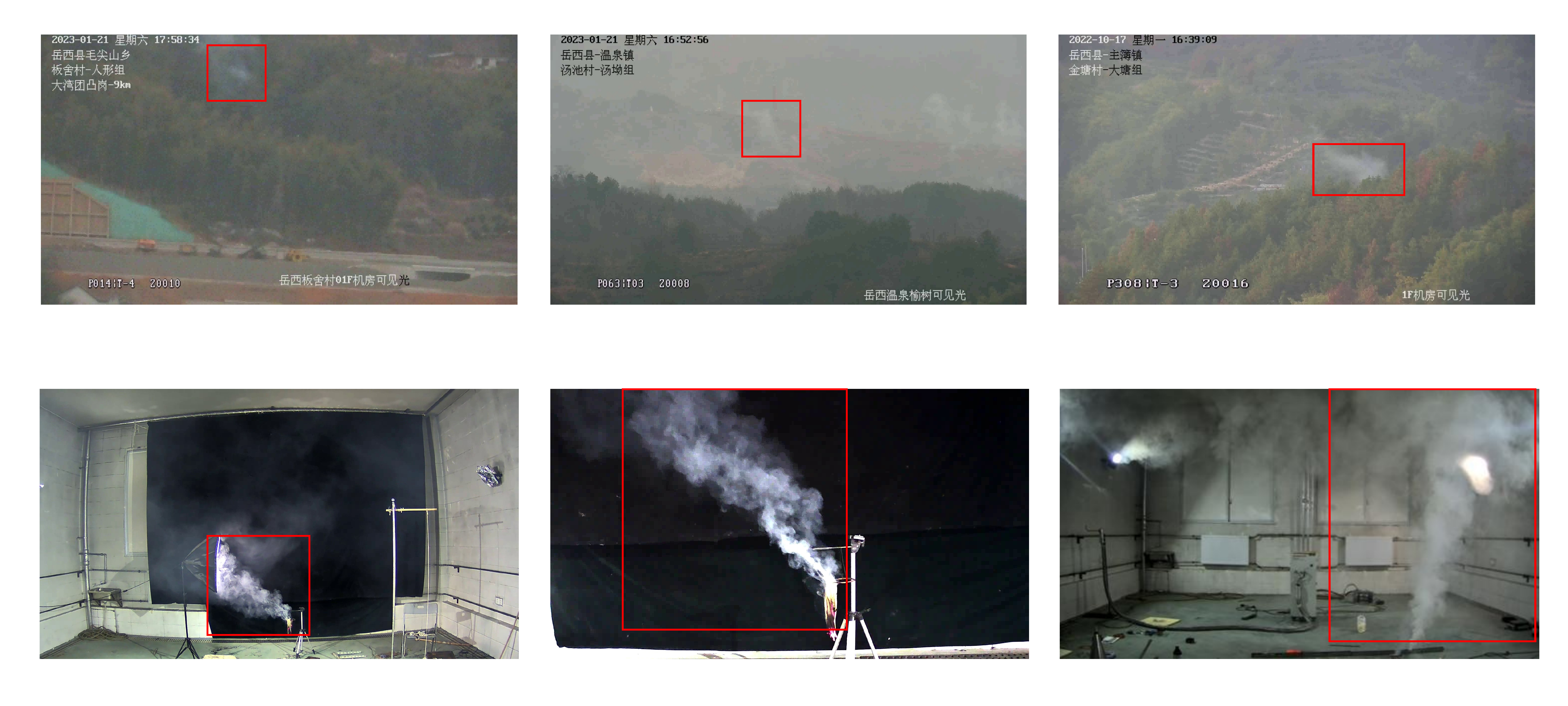}% 图片地址，可以pdf可以jpg，scale是缩放比例
        \caption{\textbf{Smoke Data}. The current smoke data generally suffer from low quality. Some of the smoke data collected outdoors are not clear enough, and their attributes are not well-defined. While smoke generated in the laboratory can be sufficiently clear, an excessive reliance on indoor smoke images can negatively impact the generalization performance of detection models.} % 图片标题
		\label{FIG:1} % 这里只要改冒号后面的数字，图片几就是几
	\end{figure*}

{
        Currently, there are two primary methods for acquiring forest fire smoke image data. The traditional approach involves collecting smoke data generated from real combustion, such as image records of actual forest fires or photographs of smoke produced during controlled ignition experiments. However, obtaining early-stage smoke images from real forest fires is extremely challenging. Controlled ignition in laboratories yields background-limited datasets, reducing diversity and generalizability. On the other hand, conducting experiments in outdoor settings, such as forests, woodlands, or nature reserves, is often impractical due to high risks and costs, making it difficult to obtain large-scale fire smoke data,and the scarcity is further exacerbated by the unpredictability of wildfire occurrences, the difficulty of capturing smoke at its initial stages, and the restrictions imposed by environmental protection and safety regulations. Due to the aforementioned reasons, the data currently used for smoke detection generally suffer from low quality, as shown in Figure \ref{FIG:1}. Therefore, relying solely on collecting smoke images from real combustion cannot fully meet the practical needs of forest fire detection. With the widespread application of digital imaging technology and deep learning, using image processing techniques or deep learning models to synthesize smoke data has become a crucial method for data acquisition. Image processing techniques primarily involve manipulating smoke in existing images through simulation or modeling software to generate synthetic images. Deep learning, on the other hand, focuses on learning features from existing images to generate new images.
    }

    \par{
        In the field of smoke and fire detection, the use of synthetic data currently relies primarily on image processing techniques and generative adversarial networks (GANs), with a smaller portion utilizing diffusion models for data synthesis. Methods based on image processing \cite{labati2013wildfire}, \cite{xu2017deep}, \cite{yuan2019wave}, \cite{mao2021wildfire}, \cite{wang2022smoke} mainly involve rendering and simulating realistic smoke and fire data using various synthesis software and modeling techniques. These approaches have been applied to early-stage smoke and fire image synthesis and integrated into detection models. Recently, some studies have employed Unreal Engine 5 \cite{wang2025fighting}, \cite{wang2024m4sfwd} to generate forest wildfire datasets, which are subsequently used for training detection models. The use of generative adversarial networks (GANs) \cite{namozov2018efficient}, \cite{huo2024enhancing} for synthesizing smoke and fire images has become a common method in recent years. These methods typically involve designing generators and discriminators for adversarial training to produce smoke and fire data similar to the dataset or to generate controllable smoke images by providing mask-based conditional inputs. Approaches based on diffusion models \cite{wang2024flame}, \cite{zheng2024fta}, \cite{zheng2024firedm} for generating smoke and fire data are still relatively few. Their design principles are similar to those of GANs, involving fine-tuning pre-trained diffusion models to generate smoke data with similar distributions or enabling controllable synthesis through editing techniques.
    }

    \par{
        Although traditional image processing techniques can generate smoke images, the fusion of the background and smoke regions is relatively simplistic, resulting in limited realism in the synthesized images. Consequently, datasets composed of such synthetic data offer only limited improvements in the accuracy and generalization of detection models. Generative Adversarial Networks (GANs), introduced by GoodFellow et al. \cite{goodfellow2014generative} in 2014, have become a pivotal technology in the field of image generation. However, GANs suffer from issues such as training instability, gradient vanishing, and mode collapse. Additionally, in the synthesis of smoke, GANs tend to produce noticeable boundaries, leading to inconsistencies in the context of the synthesized images. These challenges arise from the sensitivity of adversarial training dynamics and from the difficulty of modeling semi-transparent, fine-grained structures such as smoke while preserving global contextual consistency.Diffusion models, the state-of-the-art in image generation, were first proposed by Ho et al. \cite{ho2020denoising} in 2020. These models generate images with superior diversity and realism and allow for guided and controlled outputs through various conditions, surpassing previous image synthesis methods. The use of diffusion models for dataset augmentation has already found applications in general image editing, medical image processing, and artistic creation. Despite their impact across multiple domains, the application of diffusion models in the field of forest fire smoke remains limited. Moreover, due to the suboptimal performance of earlier diffusion-based text-to-image models in generating smoke data, which often produce erroneous and unrealistic results, and the inability of image inpainting models to achieve sufficient realism and contextual consistency in controllable smoke generation, pre-trained diffusion models cannot be directly utilized for generating forest smoke images. Furthermore, given the diversity and uncertainty inherent in diffusion model outputs, it is necessary to filter out high-quality smoke images. Therefore, there is a need to design a method specifically tailored for forest smoke generation to improve the synthesis efficiency and overall quality of smoke datasets.
        
    }
    \par{
        To address the aforementioned issues, we proposed a novel generation method tailored for smoke data, designed to generate forest smoke images with specified backgrounds and smoke contours. Additionally, we implemented a filtering process for the synthesized smoke data to enhance the quality of the dataset and improve the performance of smoke detection models.Compared with recent diffusion-based approaches, our method introduces a dedicated architecture and optimization strategy specifically designed for forest smoke generation, addressing the challenges of morphological diversity, background–foreground fusion, and controllable synthesis that previous methods have not fully resolved. Specifically, the main contributions of this paper are as follows:
    }
    \par{
        (1) We proposed an automated pipeline for generating synthetic smoke and fire datasets. By leveraging the pre-trained segmentation model SAM \cite{kirillov2023segment}, we automatically generated masks for the smoke regions based on previously annotated data used for training detection models. Additionally, we employed the multimodal model Blip2 \cite{li2023blip} to generate captions for the smoke images. After organizing and pairing these components, we obtained a dataset for training smoke image inpainting models.
    }
    \par{
        (2) We proposed a network architecture, MFGDiffusion, designed to efficiently utilize image masks and masked image information. This architecture incorporates a joint cross-attention mechanism that integrates hierarchical information from the Unet structure in Stable Diffusion. Additionally, to address the inherent priors of pre-trained diffusion inpainting models, we introduced a novel loss function, mask random difference loss, which enhances the contextual consistency between the inpainted regions and the background while improving the preservation of the image background and the diversity of the smoke in the inpainted regions.

    }
    \par{
        (3) We proposed a data filtering mechanism tailored to the characteristics of smoke. By fine-tuning an advanced multimodal large language model based on smoke properties, we acquired high-quality smoke datasets. These datasets are then automatically annotated using image masks and subsequently used to train detection models.
    }

	\par{
		The rest of this paper is organized as follows. In Section \ref{Related Work}, we provide an overview of related research relevant to this study. In Section \ref{Preliminaries}, we introduce some preliminary knowledge associated with our proposed method. Section \ref{Methodology} details the methodology we propose. In Section \ref{EXPERIMENTS}, we describe the experimental procedures and present the results. Finally, the concluding remarks are provided in the last section \ref{Conclusion}.
	}

    \begin{figure*}[h] % htbp详见说明书，记得删除括号内容
		\centering % 居中
		\includegraphics[width=0.8\linewidth]{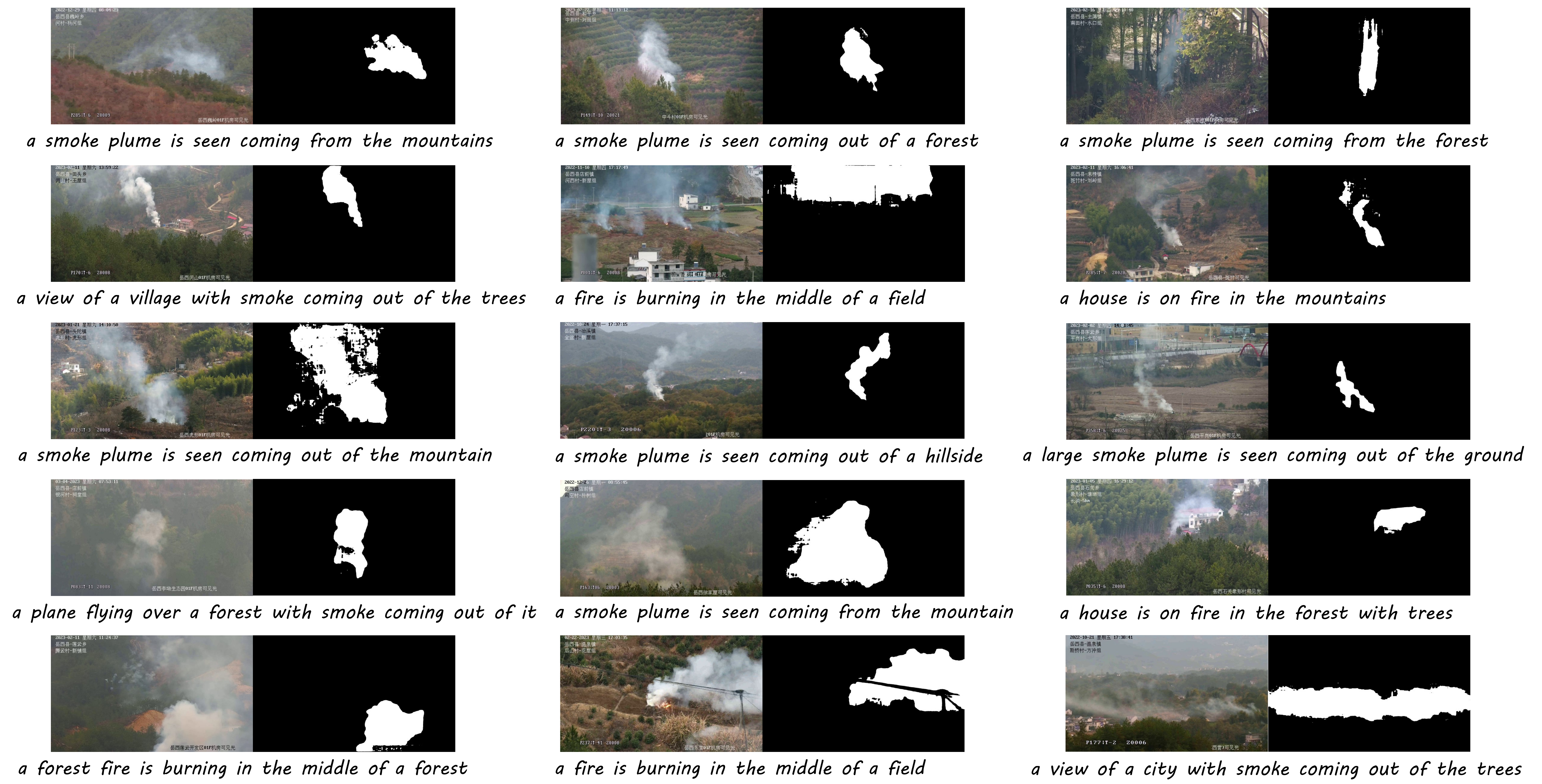}% 图片地址，可以pdf可以jpg，scale是缩放比例
        \caption{\textbf{Subset of the Dataset}. The dataset comprises source images, smoke masks, and corresponding image captions.} % 图片标题
		\label{FIG:2} % 这里只要改冒号后面的数字，图片几就是几
	\end{figure*}

\section{Related Work}
    \label{Related Work}
    \par{
		Smoke and fire image synthesis is a classic problem in the field of fire detection, aimed at enriching the training data for smoke and fire detection to enhance the performance of detection models.
	}
    \par{
        \textbf{Fire smoke synthesis}.  \cite{labati2013wildfire} synthesized wildfire smoke images using a cellular model virtual environment and image processing techniques, which were then used to train and test outdoor smoke detection algorithms.  \cite{xu2017deep} employed Blender to simulate and render smoke images and utilized domain adaptation techniques to align the feature distributions of synthetic and real smoke images, thereby improving detection performance.  \cite{zhang2018wildland} generated forest smoke images by inserting real or simulated smoke into forest backgrounds and used these synthetic images to train a Faster R-CNN model for forest fire smoke detection.  \cite{namozov2018efficient} addressed the overfitting issue caused by training networks on limited datasets by using traditional data augmentation techniques and generative adversarial networks (GANs) to increase the number of available training smoke images. Wang et al. (\cite{wang2024m4sfwd}, \cite{wang2025fighting}) enhanced the dataset by synthesizing multi-angle forest wildfire data using Unreal Engine 5 and trained their proposed detection models on this dataset.  \cite{yuan2019wave} proposed a method for synthesizing smoke to tackle the challenge of annotating ambiguous smoke images.  \cite{mao2021wildfire} simulated a large variety of richly shaped smoke using 3D modeling software and rendered virtual smoke in numerous virtual wilderness background images with diverse environmental settings, constructing a synthetic dataset to supplement smoke datasets.  \cite{wang2022smoke} improved a YOLOv5-based detection model by collecting a large number of real and synthetic smoke datasets.  \cite{huo2024enhancing} enhanced generative adversarial networks with a multi-scale fusion module, using image masks as conditions for controllable smoke image generation.  \cite{wang2024flame} proposed a diffusion model-based framework for generating wildfire image data using diffusion models and real flame image masks.  \cite{zheng2024fta} introduced a dataset quality enhancement framework based on diffusion models (DDPM) to improve the quality of low-quality fire alarm datasets.  \cite{zheng2024firedm} utilized pre-trained diffusion models to generate fire images and applied them to create high-quality fire segmentation datasets.
    }

    \par{
        Our proposed method is implemented based on the pre-trained Stable Diffusion inpainting model. Stable Diffusion, an improvement on the Latent Diffusion Model (LDM) introduced by  \cite{rombach2022high}, consists of three components: a Variational Autoencoder (VAE) \cite{kingma2013auto}, a denoising U-Net \cite{ronneberger2015u}, and a CLIP \cite{radford2021learning} text encoder. Although the pre-trained Stable Diffusion inpainting model exhibits powerful image inpainting capabilities, its performance in various derived inpainting tasks is not outstanding. Therefore, research focusing on different forms of image inpainting tasks has become a hotspot in recent years. Currently, methods for image inpainting tasks are primarily divided into two categories: those based on sampling strategy modifications and those based on dedicated inpainting models. Since End-to-End inpainting models produce more natural results in image inpainting tasks, we implement our approach based on this paradigm.
    }
    
    \par{
        \textbf{End-to-End inpainting model}. End-to-end methods typically require modifying the model or fine-tuning the original model with control conditions to learn inpainting and various editing operations in an end-to-end manner.  \cite{xie2023smartbrush} combined text and shape guidance to provide flexible control over object inpainting.  \cite{zou2024towards} proposed a method that eliminates the need for image mask inputs, leveraging the cross-modal attention capabilities of pre-trained diffusion models for instant inpainting.  \cite{zhuang2025task} improved the quality of multi-task image inpainting through task prompts and customized fine-tuning strategies.  \cite{ju2024brushnet} adopted a plug-and-play design, separating occluded image features from noise to reduce the learning burden and facilitate information fusion.  \cite{nair2024improved} proposed feature-aligned diffusion, enhancing generation quality by reducing the discrepancy between the image encoder output and the diffusion model features.  \cite{brooks2023instructpix2pix} introduced the first end-to-end image editing model based on human instructions, using large language models to generate editing instructions and datasets for training.  \cite{qi2024deadiff} incorporated the Q-former \cite{li2023blip} structure to train the model for style transfer capabilities.  \cite{wei2024omniedit} integrated multiple editing task models, generated a large number of editing pairs, and trained them on the DiT architecture, achieving unified editing functionality.
    }

% \begin{list}{}{}
% \item{\url{http://www.latex-community.org/}} 
% \item{\url{https://tex.stackexchange.com/} }
% \end{list}

\section{Preliminaries}
	\label{Preliminaries}
    \par{
		In this section, we will introduce the fundamentals of Latent Diffusion Model in Section \ref{Latent Diffusion Model}. Since end-to-end methods generally achieve better performance compared to strategy-based approaches, we also adopt an end-to-end method for our structural design in this work. Therefore, in Section \ref{Previous End-to-End Inpainting Models}, we present some basic methods currently used for end-to-end image inpainting.
	}
    \subsection{Latent Diffusion Model}
    \label{Latent Diffusion Model}
	\par{
		Latent Diffusion Models (LDMs) transform data points into noise through a series of progressive noise-adding steps and then learn a reverse process to recover the original data from the noise. In the forward process, the diffusion model gradually adds Gaussian noise $\epsilon$ to the original data $x_{0}$, causing the data distribution to progressively approach a standard normal distribution. The forward process can be described as:
	}
    \begin{equation}
        x_{t}=\sqrt{\alpha_{t}} x_{0}+\sqrt{1-\alpha_{t}} \epsilon
    \end{equation}

    \par{
        $x_{t}$ is the noised feature at step $t$ with $t \sim[1, T]$, and $\alpha$ is a hyper-parameter.
    }
    \par{
        In the backward process, given input noise $x_{t}$ sampled from a random Gaussian distribution, learnable network $\epsilon_\theta$ estimates noise at each step t conditioned on $C$. Through T progressively refining steps, $x_{0}$ is derived as the output sample:
    }
    \begin{footnotesize}
    \begin{equation}
        x_{t-1}=\frac{\sqrt{\alpha_{t-1}}}{\sqrt{\alpha_t}}x_t+\sqrt{\alpha_{t-1}}\left(\sqrt{\frac{1}{\alpha_{t-1}}-1}-\sqrt{\frac{1}{\alpha_t}-1}\right)\epsilon_\theta\left(x_t,t,C\right)
    \end{equation}
    \end{footnotesize}

    \par{
        The training of the diffusion model focuses on optimizing the denoising network $\epsilon_\theta$to perform denoising under the condition $C$ , with the objective defined as:
    }
    \begin{equation}
        \min_\theta E_{x_0,\epsilon\sim\mathcal{N}(0,I),t\sim U(1,T)}\left\|\epsilon-\epsilon_\theta\left(x_t,t,C\right)\right\|
    \end{equation}

    \par{
        The LDM architecture primarily consists of three components: a Variational Autoencoder (VAE) \cite{kingma2013auto}, which compresses images into the latent space and decodes the denoised latent representations back into images. The VAE significantly enhances the generative performance of diffusion-based methods, enabling the model to produce high-resolution images. A denoising U-Net \cite{ronneberger2015u}, which iteratively refines randomly sampled Gaussian noise through multiple U-Net layers to obtain clean latent representations. A CLIP \cite{radford2021learning} text encoder, which generates text embeddings that guide the U-Net denoising process via cross-attention mechanisms.
    }

    \subsection{Previous End-to-End Inpainting Models}
    \label{Previous End-to-End Inpainting Models}
	\par{
		Current inpainting models typically expand the input dimensions of the U-Net by concatenating the mask, masked image, and sampled Gaussian noise, which are then fed together into the U-Net for denoising. Although such end-to-end specialized inpainting models can achieve decent results through extensive training, this approach has several limitations: (1) Since the input dimensions of the U-Net are directly expanded and concatenated, the mask and masked image are influenced by noise and subsequent text conditions, preventing full utilization of these conditions to guide the denoising process. (2) We argue that the influence of the mask and masked image conditions on denoising varies across different hierarchical levels, and thus, their features should be injected separately based on specific hierarchical conditions. To address these issues, we propose MFGDiffusion to improve upon the shortcomings of previous inpainting models.
        
	}

    \begin{figure*}[h] % htbp详见说明书，记得删除括号内容
		\centering % 居中
		\includegraphics[width=0.8\linewidth]{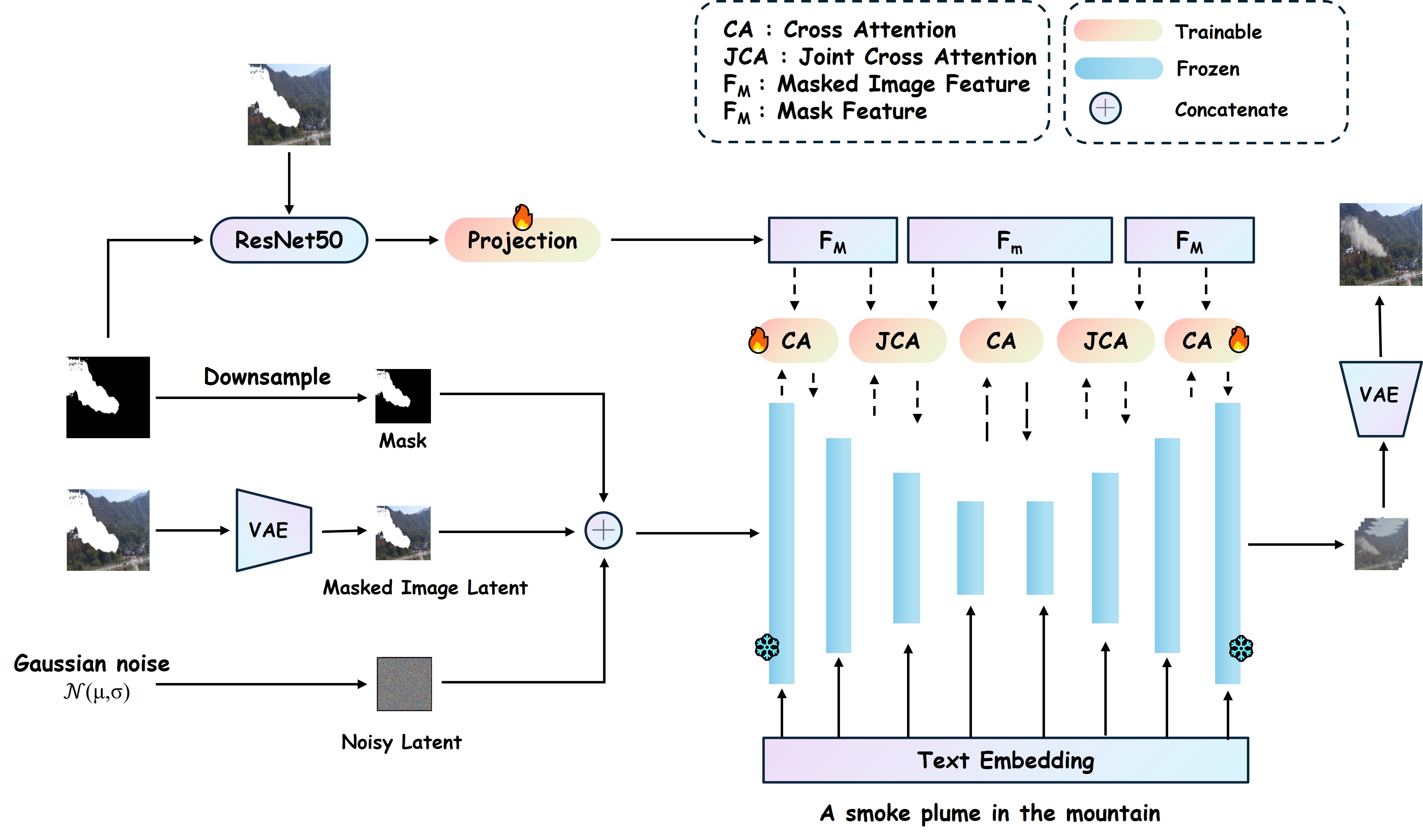}% 图片地址，可以pdf可以jpg，scale是缩放比例
        \caption{\textbf{The network architecture of MFGDiffusion}. Mask features ($F_m$) and masked image features ($F_M$) are extracted, projected, and fused with U-Net intermediate features through cross-attention ($CA$) and joint cross-attention ($JCA$). The mask mainly provides positional and morphological cues, while the masked image supplies detailed background information. Together with text embeddings, these features guide the denoising process to generate realistic smoke images. } % 图片标题
		\label{FIG:3} % 这里只要改冒号后面的数字，图片几就是几
	\end{figure*}

\section{Methodology}
	\label{Methodology}

\IEEEpubidadjcol

    \subsection{Data Acquisition}
	\par{
		The generation of smoke in a smoke-free background falls under the category of image inpainting tasks. Therefore, our dataset requires the use of data types commonly employed in image inpainting tasks: corresponding image-text pairs and masks for the smoke regions in the images. Due to the limited research in smoke and fire synthesis and the absence of publicly available datasets for smoke generation, we utilized existing smoke detection datasets to fully automate the creation of a dataset for training smoke generation models.  
	}
    
    \par{
        Specifically, to preserve the text-guided capabilities of the pre-trained image inpainting model, we employed the pre-trained image-text multimodal model BLIP-2 \cite{li2023blip} to generate captions for smoke images. BLIP-2 (Bootstrapping Language-Image Pre-training) is an advanced model based on multimodal learning, capable of understanding complex image content and generating descriptive text or answering questions about images. The captions output by the multimodal model included information about the smoke and background. However, since the target detection dataset contained detailed information such as location and time, which could affect the results of our smoke image generation, we limited the maximum number of tokens generated by the multimodal model to 20 to prevent overly detailed information from being output. Given that the smoke detection dataset included corresponding annotation files, we used the pre-trained segmentation model SAM \cite{kirillov2023segment} to segment the smoke. SAM is a general-purpose large-scale image segmentation model that can accept inputs such as positive and negative sample points, bounding boxes, and full segmentation. We employed the bounding box information obtained from previous annotation files as input to SAM, thereby obtained the mask for each image. Ultimately, we obtained over 60,000 smoke image-text pairs and corresponding mask images. These were used to train our subsequent smoke generation model. Figure \ref{FIG:2} shows a portion of the dataset.
    }

    \subsection{MFGDiffusion}
    \par{
        End-to-End image inpainting methods are typically trained based on pre-trained image inpainting models. In our proposed network architecture, we also leveraged an advanced pre-trained image inpainting model. Because the original inpainting approach struggles to fully extract feature information from the mask and masked image, we adopted SD-2-Inpainting \cite{rombach2022high} as the foundational model. To further enhance feature extraction, we introduced a pre-trained ResNet50 \cite{he2016deep} model as our feature extractor. The features output by the feature extractor are aligned with the dimensions of the intermediate representations of the U-Net through a projection layer and are fused with the intermediate representations via cross-attention layers, thereby guiding the denoising process of the U-Net. The architecture of our model is illustrated in Figure \ref{FIG:3}.
    }
    \par{
        Since both the mask and masked image serve as image condition guidance, we employed a pre-trained ResNet50 as our feature extractor. The pre-trained ResNet50 model possesses robust image analysis capabilities, enabling it to extract key features from the mask and masked image. Additionally, ResNet50, being based on a CNN architecture, shares a similar structural foundation with the denoising U-Net network. This similarity allows the image features extracted by ResNet50 to be seamlessly integrated with the intermediate representations of the denoising U-Net, and the analogous feature structure facilitates the extracted features in guiding the denoising process of the U-Net network. Specifically, since ResNet50 requires a 3-channel image as input, we used the masked image processed by the dataset but not yet reduced in dimension by the VAE as the input. For the mask, we replicated its single channel into three channels to conform to ResNet50's input requirements. The final output of ResNet50 is a 1000-class vector, but we require the features of the input image. Therefore, we took the output of the feature extractor before the pooling layer as the features of the mask and masked image, denoted as $F_m$ and $F_M$.
    }

    \begin{figure}[h] % htbp详见说明书，记得删除括号内容
		\centering % 居中
		\includegraphics[width=0.8\linewidth]{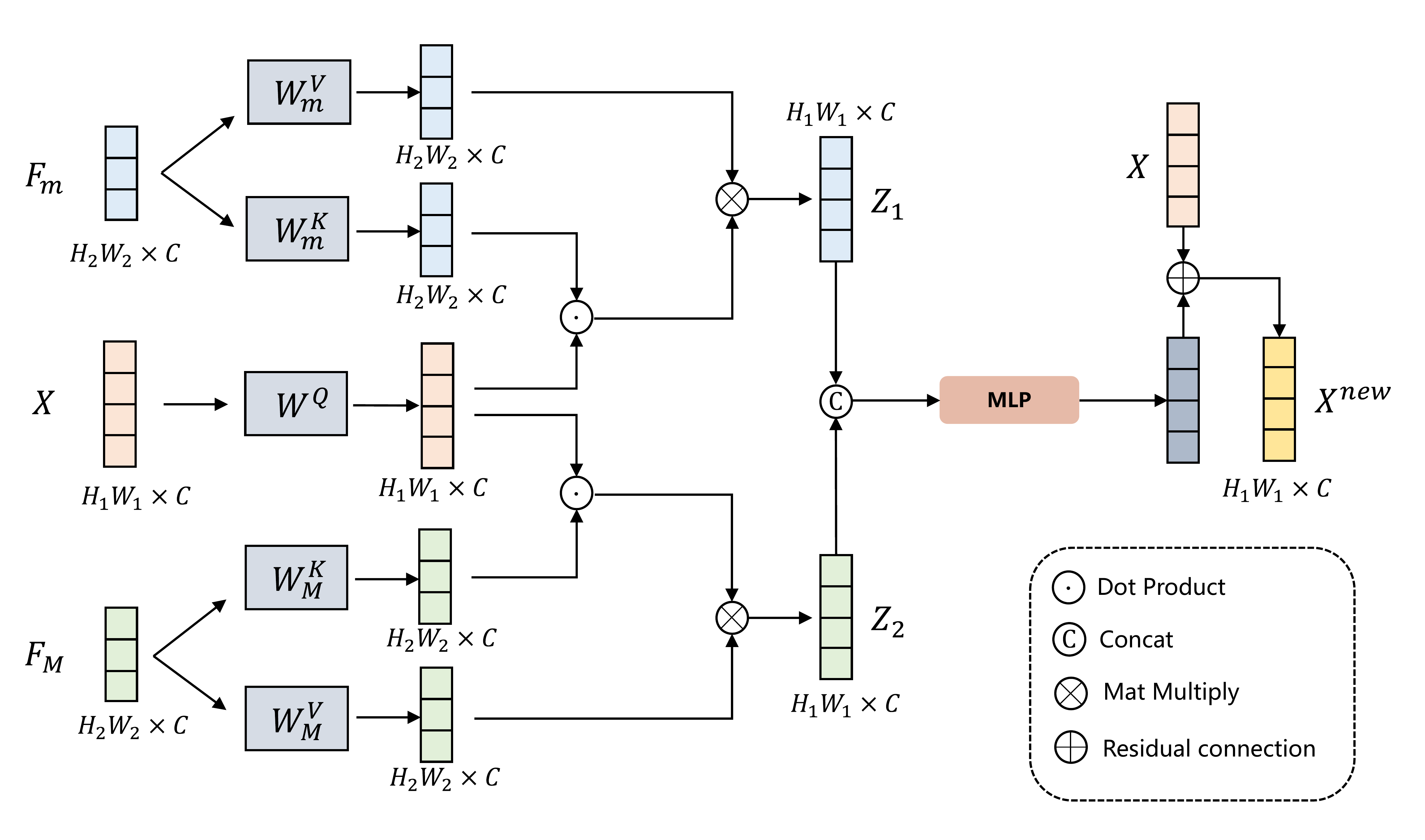}% 图片地址，可以pdf可以jpg，scale是缩放比例
        \caption{\textbf{Joint Cross-Attention for features}. Joint Cross-Attention for features. Simultaneously inject the features of both the mask and the masked image through joint cross-attention to collaboratively guide the denoising process. The two attention outputs are fused and combined with residual connections to update the U-Net representation.} % 图片标题
		\label{FIG:4} % 这里只要改冒号后面的数字，图片几就是几
	\end{figure}

    \subsection{Fusion Process}
    \par{
        The structure of the denoising U-Net is divided into an upsampling and a downsampling part, accompanied by a process from high resolution to low resolution and back to high resolution. Therefore, the U-Net is a network capable of extracting both detailed and local features of the data. To more fully utilize the features of the mask and masked image extracted by the feature extractor, we analyzed the primary roles of the mask and masked image in the image inpainting task. The mask in the image inpainting model mainly guides the location and morphological features of the image inpainting without providing detailed information about the background or the surrounding environment of the mask. Given that smoke inherently possesses diverse morphological characteristics, it does not require overly precise control of morphological information, only the relative positional information provided by the mask in the image. Therefore, we injected the mask information at the low-resolution layers of the U-Net, aiming to primarily use the positional and rough morphological information of the mask image to guide the denoising process of the U-Net. The role of the masked image in the image inpainting model is mainly to guide the generation of more detailed information such as the background and the areas surrounding the mask. Thus, we injected the features of the masked image at the high-resolution layers of the U-Net, hoping to better utilize the detailed information of the masked image to provide better background retention. Specifically, we injected only the features of the masked image at the first layer of downsampling and the last layer of upsampling in the U-Net, and only the features of the mask at the middle layers. At the intermediate layers of downsampling and upsampling, where the resolution is moderate, we injected both the features of the mask and the masked image to provide guidance at these intermediate resolution layers.
    }
    \par{
        Although both the feature extractor and the U-Net architecture are based on CNNs, the feature dimensions are $(B,C_1,H_1,W_1)$ where $B$ is the batch size, $C_1$ is the number of channels, $H_1$ is the feature map height, and $W_1$ is the feature map width,while the intermediate representations of the U-Net have dimensions $(B,C_2,H_2,W_2)$ with $C_2$ , $H_2$ , $W_2$ defined similarly. Therefore, to inject the features, we need to project the channel dimensions of the mask and masked image to match the dimensions of the U-Net's intermediate representations using a Linear layer. After projection, we used cross-attention to fuse the features of the mask and masked image with the corresponding U-Net layers and input them into the next layer. For the intermediate layers where both the mask and masked image features need to be injected simultaneously, we designed a joint cross-attention mechanism to fuse the features of both with the output of the corresponding U-Net layer, as shown in Figure \ref{FIG:4}. We first perform an attention fusion of $X$ with the mask feature $F_m$ and the masked image feature $F_M$ separately,resulting in two intermediate representations. These representations are then concatenated and passed through an MLP to enhance their fusion. In the final output, we also employed residual connections [41] to improve the effectiveness of the attention mechanism. The final output serves as the input to the next layer of the U-Net. Formally, this joint cross-attention process can be expressed as follows:
        
    }

    \begin{equation}
            Z_{\text{1}} = \operatorname{Softmax}\left(\frac{XW^Q (F_mW_m^K)^{T}}{\sqrt{d}}\right) F_mW_m^V
    \end{equation}
    
    \begin{equation}
            Z_{\text{2}} = \operatorname{Softmax}\left(\frac{XW^Q (F_MW_M^K)^{T}}{\sqrt{d}}\right) F_MW_M^V
    \end{equation}

    \begin{equation}
        X^{new} = X+\operatorname{MLP}\left(\operatorname{Concat}\left(Z_{1} , Z_{2} \right)\right)
    \end{equation}

    \par{
        $Z_{\text{1}}$ and $Z_{\text{2}}$ are the results of applying cross-attention between the features of the mask and masked image, respectively, with $X$. $W^Q$, $W_m^K$, $W_M^K$, $W_m^V$, $W_M^V$ are linear transformation matrices. MLP stands for Multi-Layer Perceptron.
    }

    \subsection{Mask Random Difference Loss}
	\par{
		Since the edges of smoke are often semi-transparent, generating smoke with semi-transparent properties within the masked area poses a significant challenge for inpainting tasks. Previous models often encountered issues such as distortion or uniformly high smoke density in both the center and edges during smoke generation, which does not align with the characteristics of early-stage forest fire smoke. Additionally, these models sometimes resulted in severe contextual inconsistencies. We argue that smoke, as a substance with diverse morphological characteristics, inherently possesses greater flexibility. In contrast, earlier inpainting models were primarily designed for objects with well-defined shapes, leading to generated objects that closely adhered to the mask. This precise generation process, guided by the mask, often resulted in contextual inconsistencies. The edges of smoke, being semi-transparent, not only reveal the smoke itself but also retain background information from before the smoke coverage. Therefore, during the mask-guided smoke generation process, it is unnecessary to strictly control the contours based on the mask edges with excessive precision.
	}
    \par
\begin{figure}[h] % htbp详见说明书，记得删除括号内容
		\centering % 居中
		\includegraphics[width=0.6\linewidth]{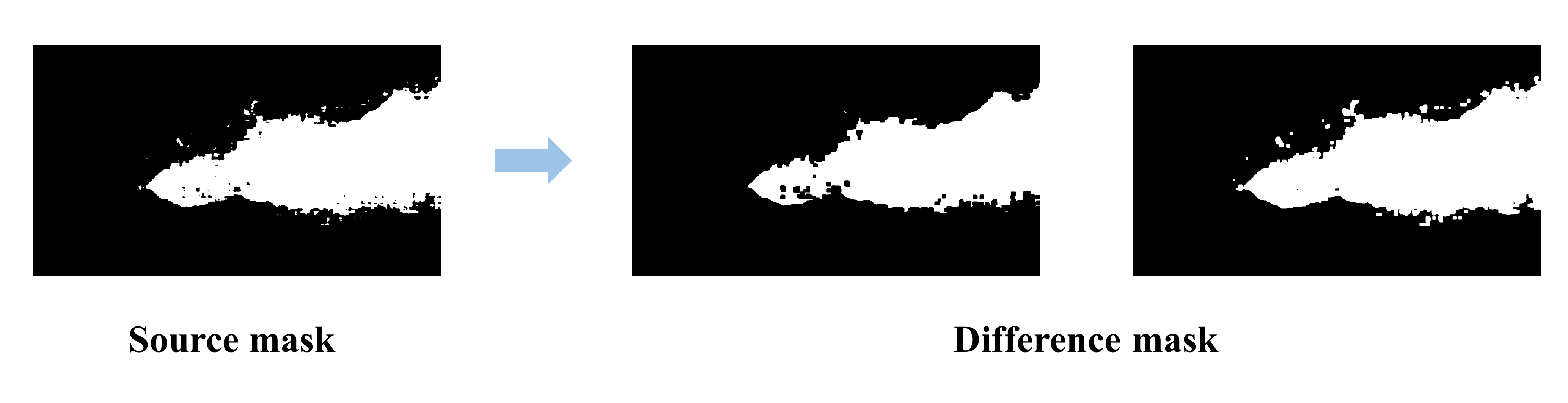}% 图片地址，可以pdf可以jpg，scale是缩放比例
        \caption{\textbf{The operation of blurring the mask}. On the left is the original mask, and on the right is the mask after random dilation and erosion. It can be observed that the mask becomes randomly denser and sparser at the edges.} % 图片标题
		\label{FIG:5} % 这里只要改冒号后面的数字，图片几就是几
	\end{figure}

{
        Inspired by the blurred characteristics of smoke edges, we propose a Mask Random Difference Loss to further enhance the quality of smoke generation, as well as the semi-transparency and diversity of smoke within the masked region. By randomly dilating and eroding the mask within a certain proportional range, we blured the boundaries of the mask and apply the blurred mask to the denoising loss. Specifically, we performed the aforementioned random dilation or erosion operations on the mask tensor $M$ using morphological operations, resulting in a difference mask $M'$. The states of the mask before and after the operation are illustrated in Figure \ref{FIG:5}. We limited the process to three rounds of morphological operations, with the kernel size for each random dilation or erosion ranging between 10 and 20. During each denoising iteration, we overlaid the randomly generated Gaussian noise and the predicted noise obtained from the denoising U-Net with $M'$, and then computed the loss. Since this loss does not rely on the precise shape of the mask for constraints, it mitigates the distortion caused by overly precise generation of smoke edges.
    }

    \par{
        We retain the original loss function and introduce the Mask Random Difference Loss with a weighting factor to optimize the training performance of the model. The total loss is formulated as follows:
    }

    \begin{footnotesize}
    \begin{equation}
        L=\omega Mse\left(M^{\prime}\epsilon,M^{\prime}\epsilon_\theta(x_t,t,C_{all})\right)+(1-\omega)Mse\left(\epsilon,\epsilon_\theta(x_t,t,C_{all})\right)
    \end{equation}
    \end{footnotesize}
    \par{
        We continued to employ Mean Squared Error (MSE) to compute the loss, where $\omega$ represents the weight of the Mask Random Difference Loss, and $C_{all}$ denotes the conditions guiding the denoising process, including text, mask, and masked image. In Section \ref{Ablation}, we designed an ablation study for the Mask Random Difference Loss, ultimately setting $\omega=0.4$ as the parameter for model training.
    }

    \subsection{Synthetic Data Filtering}
	\par{
		During the data synthesis phase, since our ultimate goal is to generate plausible smoke in smoke-free background images while leveraging the text-guided capabilities of the pre-trained model, we require captions for smoke-free background images that include descriptions of smoke to guide the smoke generation process. Here, we utilized a large language model to directly generate smoke-inclusive captions from the captions of smoke-free images. Owing to the robust capabilities of the large language model, it typically incorporates smoke-related information into the captions of smoke-free background images in a reasonable manner, such as modifying sentence structures by adding phrases like "with smoke," "a smoke plume in," or "add smoke to." Given that the scenes are forest environments with relatively simple background information, we do not need to exert excessive control over the textual details; instead, we relied on the fundamental capabilities of the text within the pre-trained model. For the generation of synthetic smoke, we used the segmentation results from SAM as the mask input, with specific details discussed in Section \ref{Implementation Details}.
	}

    \begin{figure*}[h] % htbp详见说明书，记得删除括号内容
		\centering % 居中
		\includegraphics[width=0.8\linewidth]{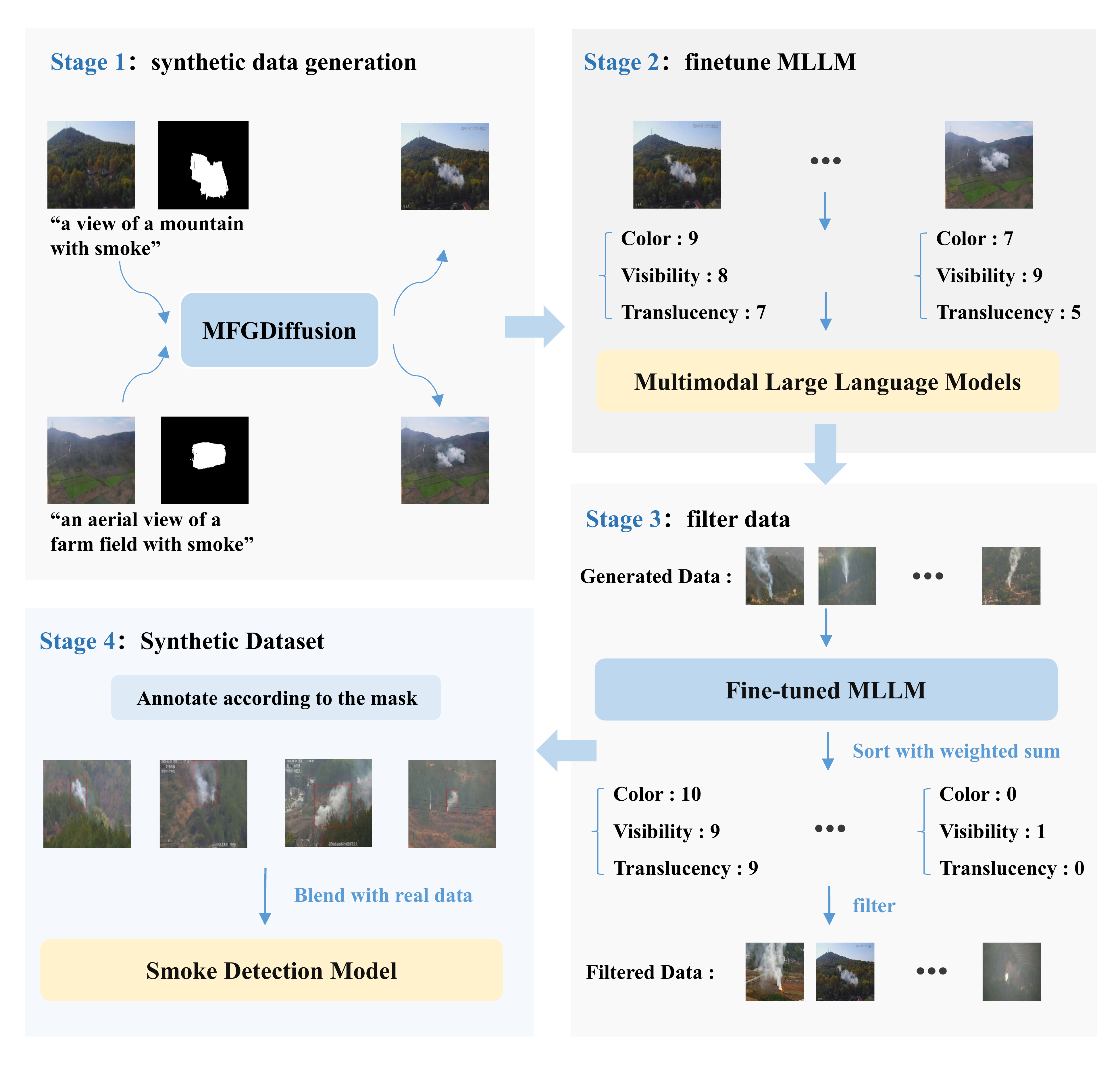}% 图片地址，可以pdf可以jpg，scale是缩放比例
        \captionsetup{justification=centering}
        \caption{\textbf{Data filtering process}. The process is divided into four stages: The first stage involves generating smoke data using random masks. The second stage entails manually annotating a small amount of data to fine-tune the multimodal large language model. The third stage utilizes the fine-tuned MLLM to evaluate three metrics of the smoke and perform weighted ranking. The fourth stage involves automatically annotating the filtered data based on the masks and combining it with real data for training the detection model.} % 图片标题
		\label{FIG:6} % 这里只要改冒号后面的数字，图片几就是几
	\end{figure*}

	\par{
		Diffusion model-based image inpainting models generally exhibit high generative diversity, and the use of random masks can also lead to inconsistencies in certain areas. These issues result in cases where the generated images may not align with the background or exhibit suboptimal quality. Therefore, to ensure the quality of the dataset for subsequent training of the smoke detection model, we need to filter the data.
	}

    \par{
        Due to the widespread application of diffusion model-based image generation techniques, many datasets for inpainting models \cite{brooks2023instructpix2pix}, \cite{qi2024deadiff}, \cite{wei2024omniedit} utilize existing image generation technologies to produce data for their specific tasks. In previous tasks involving the generation of datasets, most researchers have employed ClipScore \cite{radford2021learning} to evaluate text-image consistency, thereby filtering the generated data. However, prior work \cite{jiang2024genai} has shown that ClipScore's assessment of text-image consistency has a low correlation with human judgment. Therefore, drawing inspiration from recent work \cite{wei2024omniedit}, we utilized a multimodal large language model to evaluate the generated data. Unlike previous approaches, since the smoke data is generated from smoke-free backgrounds primarily featuring forest scenes, evaluating text-image consistency would be influenced by the similarity of the backgrounds. Thus, we utilized the multimodal large language model(MLLM) to assess the state of the generated smoke rather than text-image consistency for data filtering. Figure \ref{FIG:6} illustrates the overall workflow of our data filtering process.
    }

    \par{
        While MLLM can be directly used to evaluate the quality of generated smoke, this approach has notable limitations: the model is prone to misjudgment, and the scores tend to cluster with insufficient differentiation, making it ineffective for significantly improving the quality of the synthetic dataset. To address this, we first established a simplified evaluation framework focusing on defects in smoke generation and inherent smoke characteristics, including three metrics—color, visibility, and semi-transparency—each scored on a 0–10 scale. Testing revealed that MLLM scores still exhibited clustering tendencies. Therefore, we manually annotated 100 sets of smoke data to fine-tune Qwen2-VL and used the fine-tuned model to filter the data, ultimately selecting the top 50\% of scored samples. These selected synthetic data were combined with real data to train the detection model.
    }
    \par{
        We also discovered during experiments that using masks to generate smoke images can eliminate the need for manual annotation. By calculating the bounding box of the largest connected region within the mask, we directly obtained the annotation data for the smoke, including the length, width, and coordinates of the top-left corner of the rectangle, as illustrated in Figure \ref{FIG:6}, Stage 4. This approach significantly reduces the cost of manual annotation compared to using real data.
    }

    \begin{figure*}[h] % htbp详见说明书，记得删除括号内容
		\centering % 居中
		\includegraphics[width=0.7\linewidth]{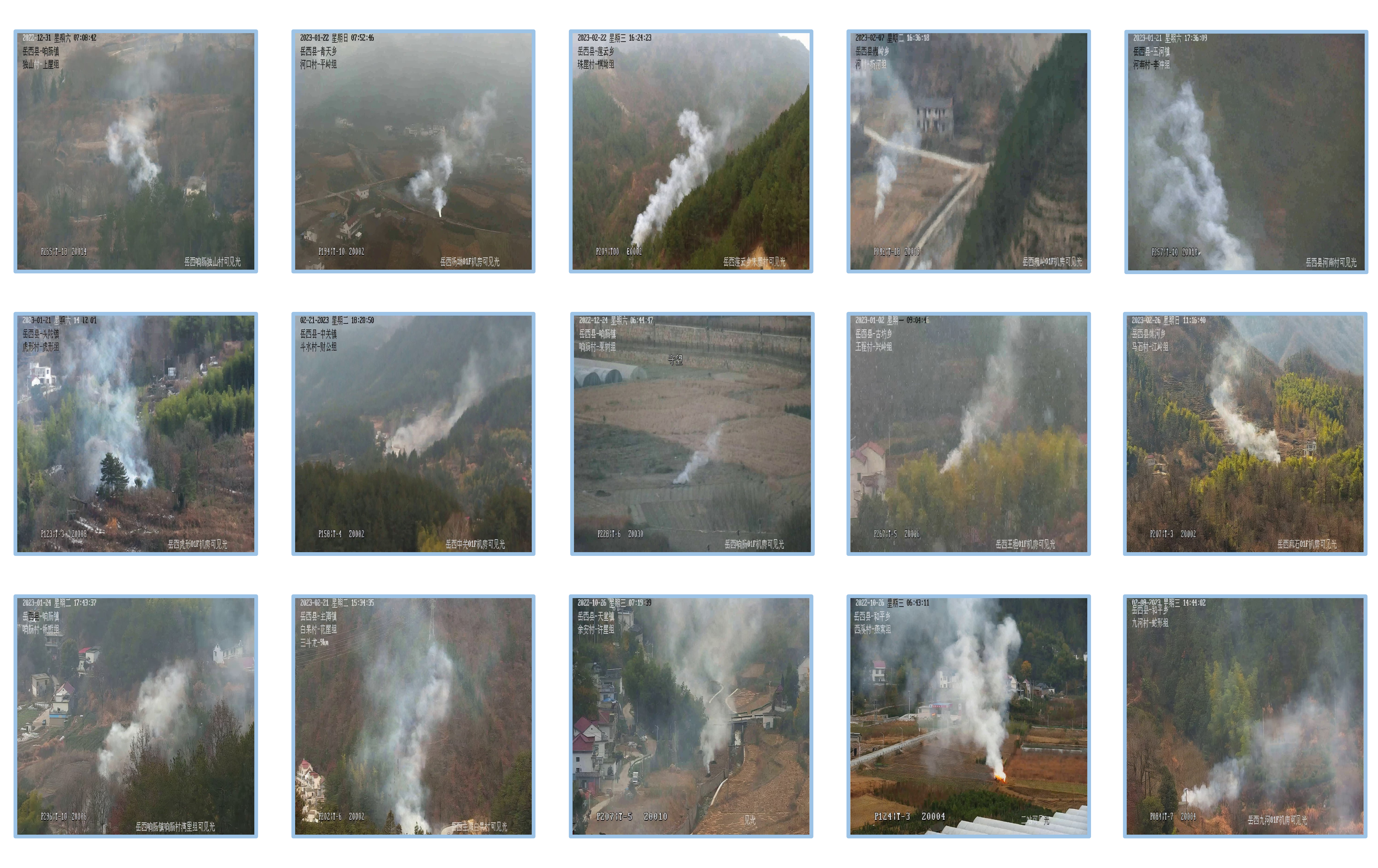}% 图片地址，可以pdf可以jpg，scale是缩放比例
        
        \caption{\textbf{A partial display of the data after filtering}. These images will be mixed with real data to jointly train the smoke detection model.} % 图片标题
		\label{FIG:7} % 这里只要改冒号后面的数字，图片几就是几
	\end{figure*}

 %% 图1  后面要加就自己复制这个改改
	\begin{figure*}[h] % htbp详见说明书，记得删除括号内容
		\centering % 居中
		\includegraphics[width=0.8\linewidth]{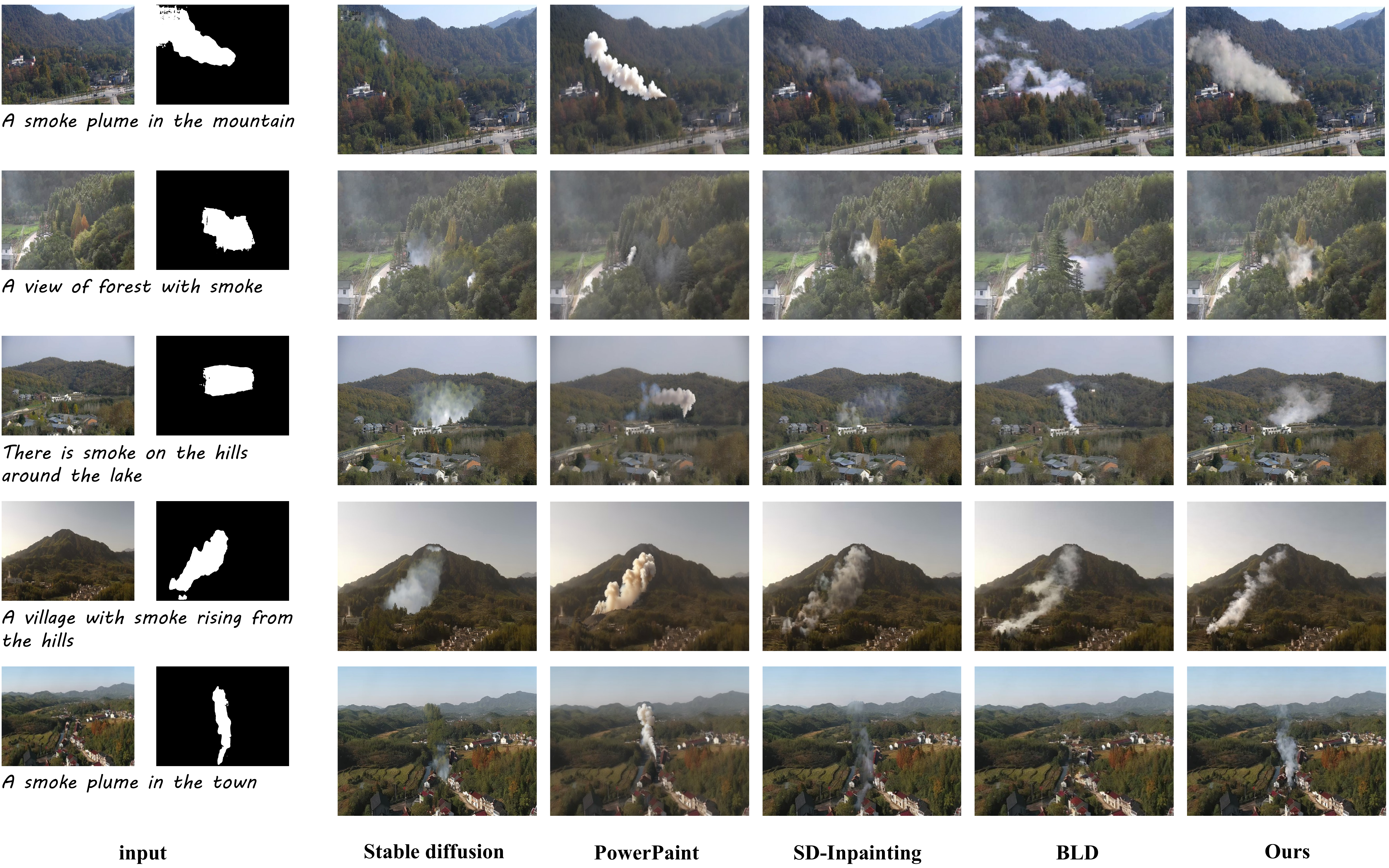}% 图片地址，可以pdf可以jpg，scale是缩放比例
        
        \caption{\textbf{Performance comparisons of MFGDiffusion and previous image inpainting methods}. Each group's input consists of a smoke-free background image, a smoke mask, and an image description. The results include five methods: (a) Stable Diffusion with strategy, (b) PowerPaint, (c) Stable Diffusion Inpainting (SD-Inpainting), (d) Blended Latent Diffusion(BLD), and (e) Ours.} % 图片标题
		\label{FIG:8} % 这里只要改冒号后面的数字，图片几就是几
	\end{figure*}

\section{EXPERIMENTS}
	\label{EXPERIMENTS}

    \subsection{Implementation Details}
    \label{Implementation Details}
    
	\par{
		\textbf{Training Details}.We employed Stable Diffusion 2 inpainting as our foundational inpainting model, which consists of a total of 4 downsampling modules, 4 upsampling modules, and an intermediate linking layer. These layers are numbered from 0 to 8, where layers 0-3 represent the downsampling process from high-resolution to low-resolution, layer 4 is the intermediate layer, and layers 5-8 represent the upsampling process from low-resolution to high-resolution. Based on the characteristics of each layer, we ultimately chose to inject the image features of the masked image into layers 0 and 8, while simultaneously injecting the features of both the mask and masked image into layers 1 and 7. The intermediate layer 4 is exclusively injected with mask features. The cross-attention modules used in each layer that receives feature injection do not share weights and independently guide the denoising process. During training, we froze the parameters of the VAE, U-Net, text encoder, and the pre-trained feature extraction module (ResNet50), updating only the projection layers after feature output and the cross-attention modules used for injecting features into the corresponding layers of the denoising U-Net. The weight for the Mask Random Difference Loss is set to 0.4. The models are trained on a single A100-80G GPU with a total batch size of 32. We adopted AdamW \cite{loshchilov2017decoupled} as the optimizer with a learning rate of $1\times10^{-4}$ and trained for a total of 20,000 iterations. For inference, we used classifier-free guidance with a guidance scale of 7.5 and perform 50 denoising steps.

	}

    \par{
		\textbf{Generate smoke data}. Prior to generating synthetic smoke data, we first needed to collect a large number of smoke-free background images to serve as the background input for smoke generation. We obtained 10,000 smoke-free background images from videos captured by cameras previously deployed for smoke detection, which provided diverse locations and perspectives. Subsequently, we processed these images using Qwen2-VL, as described earlier, to modify and augment their captions. Next, we randomly paired the masks obtained from SAM segmentation with the smoke-free background images. Since random masks may not always align well with the corresponding background images, we assigned two random masks to each smoke-free background image. This process resulted in 20,000 data pairs for smoke generation. Given that diffusion model-based implementations often exhibit high diversity in generated results, we generated three outputs for each data pair to facilitate subsequent data filtering. Ultimately, we obtained 60,000 synthetic smoke images. Among these generated results, some included mismatches between random masks and backgrounds, such as smoke generated on lake surfaces or in the sky. Therefore, we needed to filter the data to obtain a final set of high-quality synthetic smoke images.

	}

    \par{
		\textbf{Data filtering details}. We utilized the pre-trained multimodal large language model, Qwen2-VL-2B-Instruct, as our data filtering model. As described earlier, we first manually annotated 100 smoke data points, assigning scores for three specific characteristics. However, manual scoring inherently involves subjective factors, so we established certain scoring principles to ensure more objective annotations. For instance, fully transparent smoke is constrained to a score of 8-10 for semi-transparency; poorly generated smoke colors that blend with the background or appear overly white are constrained to around 5; and images where smoke generation is not clearly visible are constrained to scores of 0-2 across all three metrics. Nevertheless, due to the diversity of generated smoke data, manual scoring can generally achieve a relatively objective assessment in most cases. Through this process, we obtained evaluation data for 100 images. We then used these 100 data points to fine-tune the advanced multimodal large language model, Qwen2-VL. After fine-tuning, Qwen2-VL gained the capability to evaluate smoke color, visibility, and semi-transparency. Based on empirical knowledge of the quality of generated smoke data, we weighted the scores of three key characteristics—color, visibility, and semi-transparency—to calculate a final score, assigning weights of 0.5, 0.3, and 0.2, respectively. This weighting scheme is motivated by both domain knowledge and empirical observations. For smoke detection, color provides the most direct cue for distinguishing smoke from the background, and deviations in color severely degrade detection accuracy; therefore, it receives the highest weight. Visibility determines whether smoke can be reliably captured by the model, and insufficient visibility often leads to missed detections, making it the second most important factor. Semi-transparency, while a natural attribute of smoke, has a relatively smaller impact on model performance once color and visibility are within reasonable ranges, and is therefore assigned the lowest weight.Using Qwen2-VL, we scored the generated smoke images, performed weighted ranking, and selected the top 50\% of the synthetic smoke data as the final synthetic dataset. A partial display of our synthetic dataset is shown in Figure \ref{FIG:7}.
    
	}

    \subsection{Results}
    \label{Results}

    \par{
        \textbf{Qualitative evaluation}. We compared the most advanced inpainting models that utilize masks and text as conditions, including Stable Diffusion V2.1 \cite{rombach2022high}, which modifies strategies based on local masks, Blended Latent Diffusion (BLD) \cite{avrahami2023blended}, the end-to-end inpainting model Stable Diffusion Inpainting (SD-Inpainting), and the versatile editing model PowerPaint \cite{zhuang2025task}. To ensure fairness in qualitative comparison, we fine-tuned the base models of these methods and then applied their respective strategies or operations to generate smoke images. As shown in Figure \ref{FIG:8}, our method demonstrates superior performance compared to previous approaches. The generated smoke exhibits more reasonable results in terms of color, visibility, and semi-transparency, without issues such as excessive smoke density, distortion, or generation failure. Moreover, our method showcases strong capabilities in preserving the semi-transparent properties of smoke edges and maintaining background integrity, enabling seamless interaction with the background. In terms of generation quality, our method can produce images with a resolution of $1024\times1024$, providing the smoke detection model with richer details of both the target and the background.
    }

     \par{
        \textbf{Quantitative evaluation}. To more precisely evaluate the generative capabilities of our model and the effectiveness of the synthetic data, we employed two sets of evaluation metrics to assess our method.
    }
    \par{
        (1) Metrics for image inpainting tasks. Unlike some previous works, we did not use FID (Fréchet Inception Distance) as an evaluation metric for image quality.Unlike many unconditional generation tasks, our problem is an editing-based inpainting setting, where only the masked smoke regions are synthesized while the majority of the image remains unchanged. Therefore, the FID (Fréchet Inception Distance), which relies on global feature statistics, becomes dominated by the preserved background and cannot reliably reflect the quality of the generated smoke. Moreover, FID emphasizes overall distributional diversity, which is less relevant in our application, as the goal of forest fire smoke synthesis is to produce plausible and contextually consistent results rather than maximizing visual diversity. Instead, we adopted commonly used metrics for image generation tasks as our evaluation criteria, including: Peak Signal-to-Noise Ratio (PSNR) \cite{sheikh2006statistical}, Learned Perceptual Image Patch Similarity (LPIPS) \cite{zhang2018unreasonable}, Mean Squared Error (MSE), and Structural SIMilarity (SSIM) \cite{wang2004image}. Additionally, we employed CLIP Similarity (CLIP Sim) \cite{radford2021learning} to evaluate the similarity between the generated smoke and the text, thereby assessing the realism of the smoke and the quality of the generated images.
    }   

    % 生成实验指标对比
    \renewcommand{\arraystretch}{1.5}
    \setlength{\tabcolsep}{3mm}
    \begin{table}[ht]
    \centering
    \rmfamily
    \caption{\textbf{Comparison of generative models}. Our method achieves state-of-the-art or competitive results across various metrics for smoke image generation.}
    \label{tbl1}
    %\resizebox{.66\linewidth}{!}{%

    \resizebox{\linewidth}{!}{ 
    \begin{tabular}{@{}l|SSSS|S@{}}
        \toprule
        \textbf{Metrics} & \multicolumn{4}{c|}{\textbf{Image Quality}} & \multicolumn{1}{c}{\textbf{Text Align}} \\
        \midrule
        \textbf{Models} & {\textbf{PSNR}} & {\textbf{SSIM}} & {\textbf{LPIPS}} &{\textbf{MSE}} & {\textbf{ClipSim}} \\
        \midrule
        Stable Diffusion & {26.11} & {0.7891} & {0.095} & {188.2} & {24.94} \\
        PowerPaint & {24.14} & {0.761} & {0.243} & {267.5} & \textbf{25.78} \\
        SD-Inpainting & {25.71} & {0.7682} & {0.104} & {184.5} & {25.58} \\
        BLD & {25.88} & {0.8071} & {0.133} & {173.2} & {25.58} \\
        Ours & \textbf{28.06} & \textbf{0.8763} & \textbf{0.078} 
        & \textbf{105.3} & {25.53} \\
        \bottomrule
    \end{tabular}
    }
    %}
    \end{table}
    
    \par{
        (2) Metrics for object detection tasks. We used four metrics to evaluate the performance of models trained on both real and mixed datasets: Precision (Prec), Recall, mAP50, and mAP50-95. Precision measures the proportion of correctly predicted positive instances among all predicted positive instances. Recall measures the proportion of actual positive instances that are correctly identified. mAP50 is used to evaluate the accuracy of object localization by the model at different confidence levels, considering only predictions with an Intersection over Union (IoU) greater than or equal to 0.5. mAP50-95 is an extended version of mAP50, which not only calculates the Average Precision (AP) at an IoU threshold of 0.5 but also computes the AP at multiple IoU thresholds ranging from 0.5 to 0.95 in increments of 0.05, ultimately averaging these AP values to obtain mAP50-95. These metrics help us assess the improvement in the detection model's performance due to the synthetic data.
    }

    \begin{table*}[t]
    \centering
    \rmfamily
    \setlength{\tabcolsep}{3mm}
    \caption{\textbf{Comparison of different detection models}. By combining real data with synthetic data, we significantly improved the performance of the smoke detection model.}
    \label{tbl2}
    \begin{tabular}{@{}l|SSSS|SSSS}
        \toprule
        \textbf{DataSet} & \multicolumn{4}{c|}{\textbf{Real-world datasets}} & \multicolumn{4}{c}{\textbf{Mixed datasets}} \\
        \midrule
        \textbf{Metrics} & {\textbf{mAP50}} & {\textbf{mAP50-95}} & 
        {\textbf{Prec}} & {\textbf{Recall}} & 
        {\textbf{mAP50}} & {\textbf{mAP50-95}} &
        {\textbf{Prec}} & {\textbf{Recall}} \\
        \midrule
        Yolov6 & {0.589} & {0.298} &  {0.679} & {0.567} & \textbf{0.759} & \textbf{0.464} & \textbf{0.781} & \textbf{0.677}  \\
        Fast R-CNN & {0.657} & {0.292} &  {0.628} & {0.712} & \textbf{0.729} & \textbf{0.383} & \textbf{0.725} & \textbf{0.733}  \\
        Yolov8 & {0.635} & {0.320} &  {0.690} & {0.632} & \textbf{0.795} & \textbf{0.495} & \textbf{0.790} & \textbf{0.728}  \\
        RetinaNet& {0.678} & {0.314} &  {0.771} & {0.649} & \textbf{0.806} & \textbf{0.466} & \textbf{0.829} & \textbf{0.764}  \\
        Yolov9 & {0.654} & {0.335} &  {0.714} & {0.63} & \textbf{0.784} & \textbf{0.502} & \textbf{0.773} & \textbf{0.721}  \\
        RT-DETR1 & {0.672} & {0.311} &  {0.821} & {0.616} & \textbf{0.823} & \textbf{0.514} & \textbf{0.863} & \textbf{0.786}  \\
        Yolov10 & {0.614} & {0.295} &  {0.675} & {0.600} & \textbf{0.777} & \textbf{0.483} & \textbf{0.815} & \textbf{ 0.688}  \\
        RT-DETR2 & {0.713} & {0.336} &  {0.778} & {0.722} & \textbf{0.833} & \textbf{0.523} & \textbf{0.837} & \textbf{0.816}  \\
        Yolov11 & {0.656} & {0.332} &  {0.761} & {0.595} & \textbf{0.793} & \textbf{0.495} & \textbf{0.811} & \textbf{0.721}  \\
        Yolov12 & {0.638} & {0.301} &  {0.765} & {0.601} & \textbf{0.797} & \textbf{0.492} & \textbf{0.849} & \textbf{0.722}  \\
        \bottomrule
    \end{tabular}
    \end{table*}

    \par{
        Table \ref{tbl1} presents the comparison results between our method and previous works, demonstrating the effectiveness of MGFDiffusion in terms of image quality and text alignment. The modified Stable Diffusion, which employs a strategy-based approach, exhibits the poorest image quality. This is because the smoke in the images is generated entirely based on the mask, without considering the interaction between the smoke edges and the background, resulting in either failed smoke generation or overly dense and distorted smoke. PowerPaint performs poorly in generating semi-transparent objects, which we attribute to its training being overly biased toward generating objects with fixed shapes. Simply fine-tuning SD-Inpainting is insufficient to fully utilize the information from the mask and masked image, making it difficult to generate reasonably visible smoke within the mask and causing interference from the model's prior knowledge. Blended Latent Diffusion shows unstable generation performance, failing to produce smoke in certain scenarios.
    }
    
     \begin{table*}[t]
    \centering
    \rmfamily
    \setlength{\tabcolsep}{3mm}
    \caption{\textbf{Comparison of synthetic data and real data}. With an equivalent amount of data, the inclusion of our synthetic data results in more substantial performance gains for the smoke detection model compared to the addition of real data.}
    \label{tbl3}
    \begin{tabular}{@{}l|SSSS|SSSS}
        \toprule
        \textbf{DataSet} & \multicolumn{4}{c|}{\textbf{Real data + Real data}} & \multicolumn{4}{c}{\textbf{Real data + Synthetic data}} \\
        \midrule
        \textbf{Metrics} & {\textbf{mAP50}} & {\textbf{mAP50-95}} & 
        {\textbf{Prec}} & {\textbf{Recall}} & 
        {\textbf{mAP50}} & {\textbf{mAP50-95}} &
        {\textbf{Prec}} & {\textbf{Recall}} \\
        \midrule
        Yolov6 & {0.637} & {0.330} &  {0.641} & {0.642} & \textbf{0.759} & \textbf{0.464} & \textbf{0.781} & \textbf{0.677}  \\
        Fast R-CNN & {0.683} & {0.304} &  {0.602} & {0.755} & \textbf{0.729} & \textbf{0.383} & \textbf{0.725} & \textbf{0.733}  \\
        Yolov8 & {0.675} & {0.350} &  {0.710} & {0.665} & \textbf{0.795} & \textbf{0.495} & \textbf{0.790} & \textbf{0.728}  \\
        RetinaNet& {0.710} & {0.323} &  {0.723} & {0.707} & \textbf{0.806} & \textbf{0.466} & \textbf{0.829} & \textbf{0.764}  \\
        Yolov9 & {0.673} & {0.345} &  {0.712} & {0.656} & \textbf{0.784} & \textbf{0.502} & \textbf{0.773} & \textbf{0.721}  \\
        RT-DETR1 & {0.702} & {0.337} &  {0.758} & {0.690} & \textbf{0.823} & \textbf{0.514} & \textbf{0.863} & \textbf{0.786}  \\
        Yolov10 & {0.670} & {0.341} &  {0.737} & {0.636} & \textbf{0.777} & \textbf{0.483} & \textbf{0.815} & \textbf{ 0.688}  \\
        RT-DETR2 & {0.742} & {0.370} &  {0.752} & {0.745} & \textbf{0.833} & \textbf{0.523} & \textbf{0.837} & \textbf{0.816}  \\
        Yolov11 & {0.680} & {0.349} &  {0.647} & {0.725} & \textbf{0.793} & \textbf{0.495} & \textbf{0.811} & \textbf{0.721}  \\
        Yolov12 & {0.679} & {0.361} &  {0.781} & {0.642} & \textbf{0.797} & \textbf{0.492} & \textbf{0.849} & \textbf{0.722}  \\
        \bottomrule
    \end{tabular}
    \end{table*}

    \begin{figure}[h] % htbp详见说明书，记得删除括号内容
		\centering % 居中
		\includegraphics[width=0.8\linewidth]{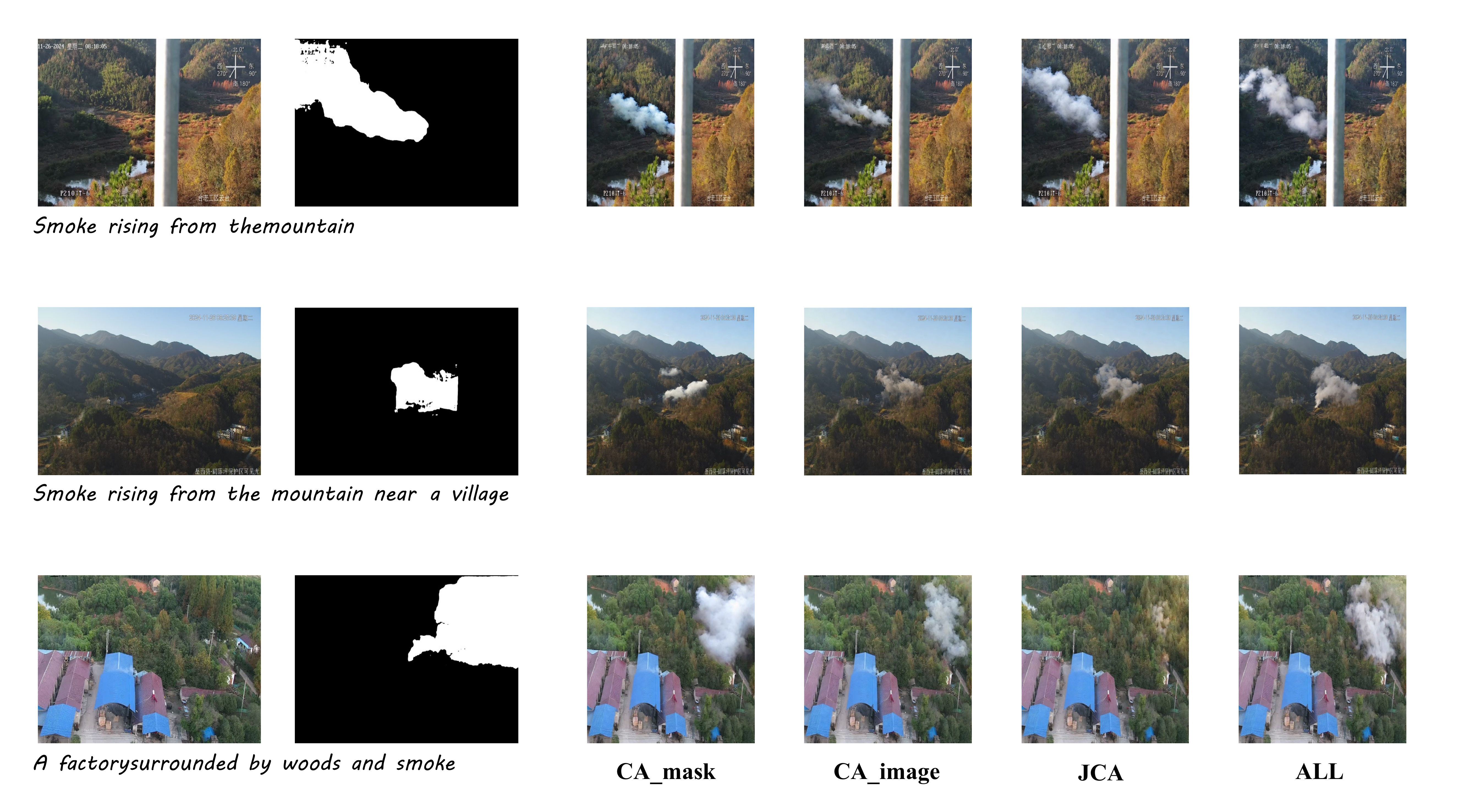}% 图片地址，可以pdf可以jpg，scale是缩放比例
        
        \caption{\textbf{Effect of modules in network architecture}. Using only the mask can lead to a decrease in the ability to interact with the background, while using only the masked image or not employing the Joint Cross-Attention (JCA) mechanism can result in varying degrees of degradation in the generation capability.} % 图片标题
		\label{FIG:9} % 这里只要改冒号后面的数字，图片几就是几
	\end{figure}

    \par{
        In Table \ref{tbl2}, we demonstrated the effectiveness of our synthetic dataset. We conducted experiments using widely adopted detection models, including Yolov6, Yolov8, Yolov9, Yolov10, and Yolov11. To evaluate the performance of the trained detection models in real-world smoke detection scenarios, we used only real data as the test set. The experimental results indicate that mixing the synthetic dataset with the real dataset leads to significant improvements across all metrics for each detection model. This confirms that our high-quality synthetic dataset can effectively enhance the performance of smoke detection models in real-world environments.
    }
    \par{
        We also compared the effects of replacing the synthetic data with real data during model training, as shown in Table \ref{tbl3}. It can be observed that, with the same amount of data, incorporating synthetic data leads to greater improvements in the smoke detection model compared to adding real data. We attributed this to the fact that the synthetic smoke data is inherently very close to real data, and through the data filtering process, only high-quality synthetic data is retained. Therefore, adding this portion of high-quality synthetic data to the real dataset yields better results than simply adding more real data.
    }

    \begin{table*}[t]
    \centering
    \rmfamily
    \setlength{\tabcolsep}{3mm}
    \caption{\textbf{Effect of filtering module}. The filtering module significantly enhances data quality, and utilizing this high-quality data for training leads to a smoke detection model with superior performance.}
    \label{tbl4}
    \begin{tabular}{@{}l|SSSS|SSSS}
        \toprule
        \textbf{DataSet} & \multicolumn{4}{c|}{\textbf{Random datasets}} & \multicolumn{4}{c}{\textbf{Filtered datasets}} \\
        \midrule
        \textbf{Metrics} & {\textbf{mAP50}} & {\textbf{mAP50-95}} & 
        {\textbf{Prec}} & {\textbf{Recall}} & 
        {\textbf{mAP50}} & {\textbf{mAP50-95}} &
        {\textbf{Prec}} & {\textbf{Recall}} \\
        \midrule
        Yolov6 & {0.695} & {0.403} &  {0.704} & {0.666} & \textbf{0.759} & \textbf{0.464} & \textbf{0.781} & \textbf{0.677}  \\
        Fast R-CNN & {0.681} & {0.335} &  {0.661} & {0.705} & \textbf{0.729} & \textbf{0.383} & \textbf{0.725} & \textbf{0.733}  \\
        Yolov8 & {0.752} & {0.438} &  {0.781} & {0.688} & \textbf{0.795} & \textbf{0.495} & \textbf{0.79} & \textbf{0.728}  \\
        RetinaNet& {0.771} & {0.426} &  {0.782} & {0.735} & \textbf{0.806} & \textbf{0.466} & \textbf{0.829} & \textbf{0.764}  \\
        Yolov9 & {0.736} & {0.448} &  {0.732} & {0.681} & \textbf{0.784} & \textbf{0.502} & \textbf{0.773} & \textbf{0.721}  \\
        RT-DETR1 & {0.786} & {0.461} &  {0.839} & {0.754} & \textbf{0.823} & \textbf{0.514} & \textbf{0.863} & \textbf{0.786}  \\
        Yolov10 & {0.726} & {0.417} &  {0.792} & {0.655} & \textbf{0.777} & \textbf{0.483} & \textbf{0.815} & \textbf{ 0.688}  \\
        RT-DETR2 & {0.799} & {0.476} &  {0.805} & {0.777} & \textbf{0.833} & \textbf{0.523} & \textbf{0.837} & \textbf{0.816}  \\
        Yolov11 & {0.745} & {0.441} &  {0.776} & {0.692} & \textbf{0.793} & \textbf{0.495} & \textbf{0.811} & \textbf{0.721}  \\
        Yolov12 & {0.756} & {0.438} &  {0.832} & {0.675} & \textbf{0.797} & \textbf{0.492} & \textbf{0.849} & \textbf{0.722}  \\
        \bottomrule
    \end{tabular}
    \end{table*}

    \subsection{Ablation Study}
    \label{Ablation}

    \par{
        In the proposed model architecture, the mask and masked image exhibit distinct guiding effects on the denoising process of the U-Net. In the injection module, the JCA (Joint Contextual Attention) mechanism is employed to simultaneously inject features from both the mask and masked image. We conducted multiple sets of experiments to validate the effectiveness of this module. Through these experiments, we observed that incorporating the JCA and injecting features from the mask and masked image at different levels significantly enhances the smoke generation performance. The experimental comparison is illustrated in Figure \ref{FIG:9}, where CA\_mask denotes the configuration without the JCA module and only injects features from the mask, while CA\_masked image represents the configuration without the JCA module and only injects features from the masked image. JCA refers to the configuration where the JCA module is removed, but the distinct injection of features is still maintained at other injection layers.
    }

    \begin{figure}[h] % htbp详见说明书，记得删除括号内容
		\centering % 居中
		\includegraphics[width=0.7\linewidth]{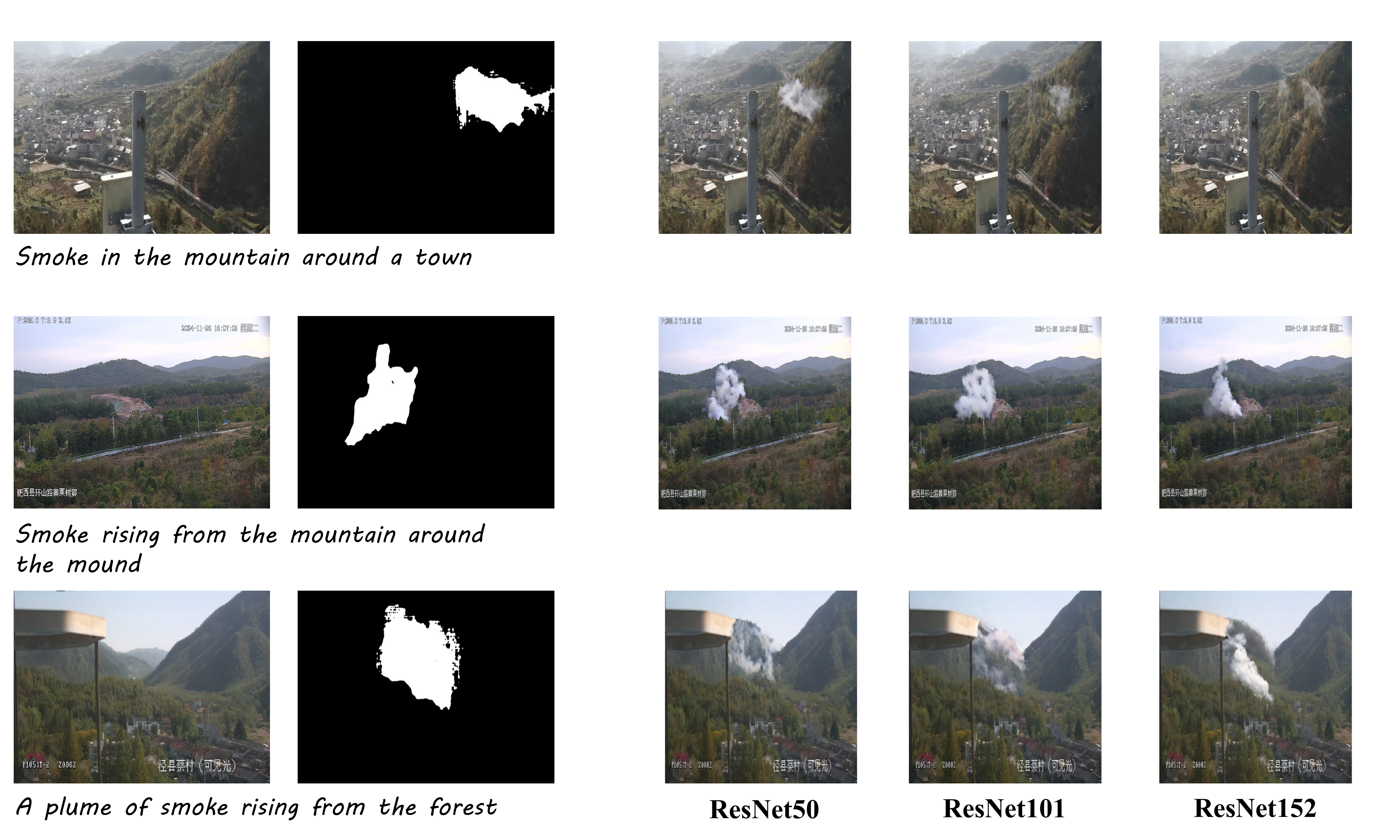}% 图片地址，可以pdf可以jpg，scale是缩放比例
        
        \caption{\textbf{Effect of feature extractor}. ResNet50 serves as a more stable feature extractor and yields better generation outcomes.} % 图片标题
		\label{FIG:10} % 这里只要改冒号后面的数字，图片几就是几
	\end{figure}

    \par{
        In our study, we initially selected ResNet50 as the feature extractor. To investigate whether larger-scale feature extractors could yield superior results, we conducted experiments using larger models from the ResNet series, namely ResNet101 and ResNet152. The results are illustrated in Figure \ref{FIG:10}. Through these experiments, we observed that employing larger ResNet models did not enhance the quality of smoke generation; instead, it led to incomplete or even failed smoke generation in some cases. Therefore, to ensure the quality and stability of the generated smoke, we consistently used ResNet50 as the feature extractor throughout this work.
    }

    \begin{figure}[h] % htbp详见说明书，记得删除括号内容
		\centering % 居中
		\includegraphics[width=0.8\linewidth]{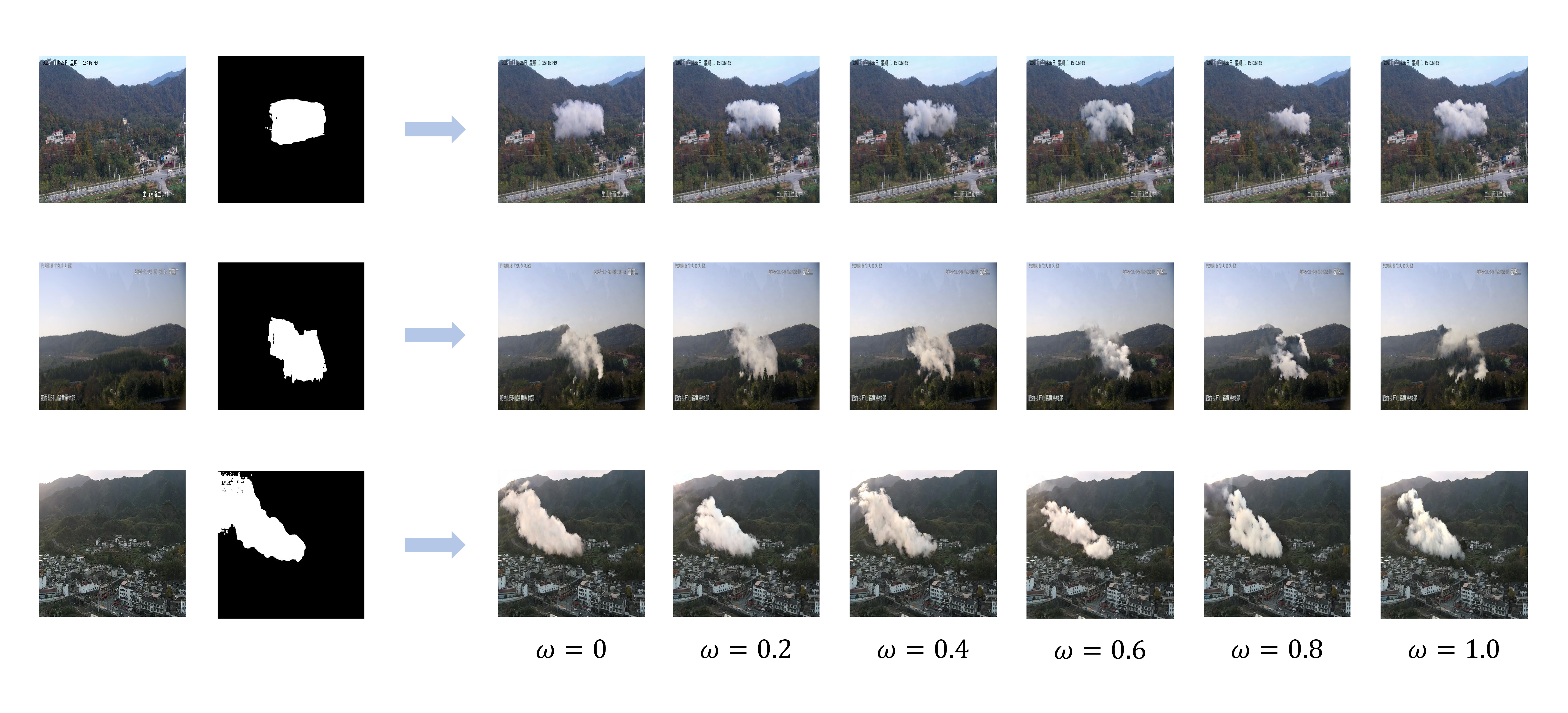}% 图片地址，可以pdf可以jpg，scale是缩放比例
        
        \caption{\textbf{Effect of mask random difference loss}. Varying the weights can significantly influence the model's capability to effectively interact with the background.} % 图片标题
		\label{FIG:11} % 这里只要改冒号后面的数字，图片几就是几
	\end{figure}

     \par{
        To validate the effectiveness of the proposed Mask Random Difference Loss in model training, we conducted ablation experiments to investigate the impact of different weights for the Mask Random Difference Loss on the quality of smoke generation. Figure \ref{FIG:11} illustrates the smoke generation results obtained when the Mask Random Difference Loss is incorporated into model training with varying weights. The experiments were conducted using the same seed for smoke generation. We observed that the smaller the weight of the Mask Random Difference Loss, the less correlated the generated smoke is with the mask edges. However, if the Mask Random Difference Loss is used as the primary loss for training, the control of the mask over the model significantly diminishes. For instance, as shown in Figure \ref{FIG:11}, when the Mask Random Difference Loss is the sole loss $\omega = 1.0$, the generated results no longer adhere to the mask shape. Therefore, for the synthetic dataset, we set the weight of the Mask Random Difference Loss to 0.4 to balance these two properties.
    }

    \par{
        We also designed experiments to validate the effectiveness of the multimodal large language model in data filtering. For this purpose, we created two datasets: one using filtered synthetic smoke data and the other using randomly selected synthetic smoke data, both mixed with real data in a 1:1 ratio, with a positive-to-negative sample ratio of 1:1. We trained multiple detection models on these two datasets separately. As shown in Table \ref{tbl4}, significant improvements were observed across all four metrics used in our evaluation for different detection models. 
    }

\section{Conclusion}
	\label{Conclusion}

    \par{
		This paper proposes a comprehensive pipeline for forest fire smoke image generation and detection, which includes the acquisition of a pre-trained diffusion model dataset, a feature extraction module, a novel Mask Random Difference loss function, and a data filtering module based on a Multimodal Large Language Model. By leveraging multiple pre-trained models, we generated a training dataset of over 60,000 samples and trained the model with a newly designed projection layer and cross-attention module. Experiments demonstrate that the proposed MFGDiffusion architecture can generate realistic smoke images, outperforming existing methods in visual aspects such as smoke texture, morphology, and transparency, while also achieving better performance on metrics including PSNR, SSIM, LPIPS, MSE, and ClipSim.
	}
    \par{
        Furthermore, based on the characteristics of the generated smoke, we fine-tuned a Multimodal Large Language Model to filter high-quality datasets. By combining synthetic data with real data, we effectively alleviated the issues of limited availability and insufficient diversity in smoke datasets, significantly improving the performance of smoke detection models. 
    }
    \par{ Although the method effectively improves the realism of smoke generation and the performance of detection models, it has not been extended to the video modality and cannot be used for video-based detection models. Our future work will focus on research into generating natural smoke videos, with particular attention to the fidelity and physical rationality during the smoke generation process.
    }

\bibliographystyle{IEEEtran}
\bibliography{IEEEabrv,reference}

\begin{thebibliography}{10}
\providecommand{\url}[1]{#1}
\csname url@samestyle\endcsname
\providecommand{\newblock}{\relax}
\providecommand{\bibinfo}[2]{#2}
\providecommand{\BIBentrySTDinterwordspacing}{\spaceskip=0pt\relax}
\providecommand{\BIBentryALTinterwordstretchfactor}{4}
\providecommand{\BIBentryALTinterwordspacing}{\spaceskip=\fontdimen2\font plus
\BIBentryALTinterwordstretchfactor\fontdimen3\font minus \fontdimen4\font\relax}
\providecommand{\BIBforeignlanguage}[2]{{%
\expandafter\ifx\csname l@#1\endcsname\relax
\typeout{** WARNING: IEEEtran.bst: No hyphenation pattern has been}%
\typeout{** loaded for the language `#1'. Using the pattern for}%
\typeout{** the default language instead.}%
\else
\language=\csname l@#1\endcsname
\fi
#2}}
\providecommand{\BIBdecl}{\relax}
\BIBdecl

\bibitem{li20183d}
X.~Li, Z.~Chen, Q.~J. Wu, and C.~Liu, ``3d parallel fully convolutional networks for real-time video wildfire smoke detection,'' \emph{IEEE Transactions on Circuits and Systems for Video Technology}, vol.~30, no.~1, pp. 89--103, 2018.

\bibitem{tao2019smoke}
H.~Tao and X.~Lu, ``Smoke vehicle detection based on spatiotemporal bag-of-features and professional convolutional neural network,'' \emph{IEEE Transactions on Circuits and Systems for Video Technology}, vol.~30, no.~10, pp. 3301--3316, 2019.

\bibitem{cao2021effnet}
Y.~Cao, Q.~Tang, X.~Wu, and X.~Lu, ``Effnet: Enhanced feature foreground network for video smoke source prediction and detection,'' \emph{IEEE Transactions on Circuits and Systems for Video Technology}, vol.~32, no.~4, pp. 1820--1833, 2021.

\bibitem{jing2023smokepose}
T.~Jing, M.~Zeng, and Q.-H. Meng, ``Smokepose: End-to-end smoke keypoint detection,'' \emph{IEEE Transactions on Circuits and Systems for Video Technology}, vol.~33, no.~10, pp. 5778--5789, 2023.

\bibitem{wang2024dpmnet}
G.~Wang, H.~Li, V.~Sheng, Y.~Ma, H.~Ding, and H.~Zhao, ``Dpmnet: A remote sensing forest fire real-time detection network driven by dual pathways and multidimensional interactions of features,'' \emph{IEEE Transactions on Circuits and Systems for Video Technology}, 2024.

\bibitem{labati2013wildfire}
R.~D. Labati, A.~Genovese, V.~Piuri, and F.~Scotti, ``Wildfire smoke detection using computational intelligence techniques enhanced with synthetic smoke plume generation,'' \emph{IEEE Transactions on Systems, Man, and Cybernetics: Systems}, vol.~43, no.~4, pp. 1003--1012, 2013.

\bibitem{xu2017deep}
G.~Xu, Y.~Zhang, Q.~Zhang, G.~Lin, and J.~Wang, ``Deep domain adaptation based video smoke detection using synthetic smoke images,'' \emph{Fire safety journal}, vol.~93, pp. 53--59, 2017.

\bibitem{yuan2019wave}
F.~Yuan, L.~Zhang, X.~Xia, Q.~Huang, and X.~Li, ``A wave-shaped deep neural network for smoke density estimation,'' \emph{IEEE transactions on image processing}, vol.~29, pp. 2301--2313, 2019.

\bibitem{mao2021wildfire}
J.~Mao, C.~Zheng, J.~Yin, Y.~Tian, and W.~Cui, ``Wildfire smoke classification based on synthetic images and pixel-and feature-level domain adaptation,'' \emph{Sensors}, vol.~21, no.~23, p. 7785, 2021.

\bibitem{wang2022smoke}
Z.~Wang, L.~Wu, T.~Li, and P.~Shi, ``A smoke detection model based on improved yolov5,'' \emph{Mathematics}, vol.~10, no.~7, p. 1190, 2022.

\bibitem{wang2025fighting}
G.~Wang, H.~Li, Q.~Xiao, P.~Yu, Z.~Ding, Z.~Wang, and S.~Xie, ``Fighting against forest fire: A lightweight real-time detection approach for forest fire based on synthetic images,'' \emph{Expert Systems with Applications}, vol. 262, p. 125620, 2025.

\bibitem{wang2024m4sfwd}
G.~Wang, H.~Li, P.~Li, X.~Lang, Y.~Feng, Z.~Ding, and S.~Xie, ``M4sfwd: A multi-faceted synthetic dataset for remote sensing forest wildfires detection,'' \emph{Expert Systems with Applications}, vol. 248, p. 123489, 2024.

\bibitem{namozov2018efficient}
A.~Namozov and Y.~Im~Cho, ``An efficient deep learning algorithm for fire and smoke detection with limited data,'' \emph{Advances in Electrical and Computer Engineering}, vol.~18, no.~4, pp. 121--128, 2018.

\bibitem{huo2024enhancing}
Y.~Huo, Q.~Zhang, C.~Wang, H.~Wang, and Y.~Zhang, ``Enhancing wildfire detection: a novel algorithm for controllable generation of wildfire smoke images,'' \emph{International Journal of Wildland Fire}, vol.~33, no.~11, 2024.

\bibitem{wang2024flame}
H.~Wang, S.~P.~H. Boroujeni, X.~Chen, A.~Bastola, H.~Li, W.~Zhu, and A.~Razi, ``Flame diffuser: Wildfire image synthesis using mask guided diffusion,'' \emph{arXiv preprint arXiv:2403.03463}, 2024.

\bibitem{zheng2024fta}
H.~Zheng, G.~Wang, D.~Xiao, H.~Liu, and X.~Hu, ``Fta-detr: An efficient and precise fire detection framework based on an end-to-end architecture applicable to embedded platforms,'' \emph{Expert Systems with Applications}, vol. 248, p. 123394, 2024.

\bibitem{zheng2024firedm}
H.~Zheng, M.~Wang, Z.~Wang, and X.~Huang, ``Firedm: A weakly-supervised approach for massive generation of multi-scale and multi-scene fire segmentation datasets,'' \emph{Knowledge-Based Systems}, vol. 290, p. 111547, 2024.

\bibitem{goodfellow2014generative}
I.~Goodfellow, J.~Pouget-Abadie, M.~Mirza, B.~Xu, D.~Warde-Farley, S.~Ozair, A.~Courville, and Y.~Bengio, ``Generative adversarial nets,'' \emph{Advances in neural information processing systems}, vol.~27, 2014.

\bibitem{ho2020denoising}
J.~Ho, A.~Jain, and P.~Abbeel, ``Denoising diffusion probabilistic models,'' \emph{Advances in neural information processing systems}, vol.~33, pp. 6840--6851, 2020.

\bibitem{kirillov2023segment}
A.~Kirillov, E.~Mintun, N.~Ravi, H.~Mao, C.~Rolland, L.~Gustafson, T.~Xiao, S.~Whitehead, A.~C. Berg, W.-Y. Lo \emph{et~al.}, ``Segment anything,'' in \emph{Proceedings of the IEEE/CVF International Conference on Computer Vision}, 2023, pp. 4015--4026.

\bibitem{li2023blip}
J.~Li, D.~Li, S.~Savarese, and S.~Hoi, ``Blip-2: Bootstrapping language-image pre-training with frozen image encoders and large language models,'' in \emph{International conference on machine learning}.\hskip 1em plus 0.5em minus 0.4em\relax PMLR, 2023, pp. 19\,730--19\,742.

\bibitem{zhang2018wildland}
Q.-x. Zhang, G.-h. Lin, Y.-m. Zhang, G.~Xu, and J.-j. Wang, ``Wildland forest fire smoke detection based on faster r-cnn using synthetic smoke images,'' \emph{Procedia engineering}, vol. 211, pp. 441--446, 2018.

\bibitem{rombach2022high}
R.~Rombach, A.~Blattmann, D.~Lorenz, P.~Esser, and B.~Ommer, ``High-resolution image synthesis with latent diffusion models,'' in \emph{Proceedings of the IEEE/CVF conference on computer vision and pattern recognition}, 2022, pp. 10\,684--10\,695.

\bibitem{kingma2013auto}
D.~P. Kingma, ``Auto-encoding variational bayes,'' \emph{arXiv preprint arXiv:1312.6114}, 2013.

\bibitem{ronneberger2015u}
O.~Ronneberger, P.~Fischer, and T.~Brox, ``U-net: Convolutional networks for biomedical image segmentation,'' in \emph{Medical image computing and computer-assisted intervention--MICCAI 2015: 18th international conference, Munich, Germany, October 5-9, 2015, proceedings, part III 18}.\hskip 1em plus 0.5em minus 0.4em\relax Springer, 2015, pp. 234--241.

\bibitem{radford2021learning}
A.~Radford, J.~W. Kim, C.~Hallacy, A.~Ramesh, G.~Goh, S.~Agarwal, G.~Sastry, A.~Askell, P.~Mishkin, J.~Clark \emph{et~al.}, ``Learning transferable visual models from natural language supervision,'' in \emph{International conference on machine learning}.\hskip 1em plus 0.5em minus 0.4em\relax PMLR, 2021, pp. 8748--8763.

\bibitem{xie2023smartbrush}
S.~Xie, Z.~Zhang, Z.~Lin, T.~Hinz, and K.~Zhang, ``Smartbrush: Text and shape guided object inpainting with diffusion model,'' in \emph{Proceedings of the IEEE/CVF conference on computer vision and pattern recognition}, 2023, pp. 22\,428--22\,437.

\bibitem{zou2024towards}
S.~Zou, J.~Tang, Y.~Zhou, J.~He, C.~Zhao, R.~Zhang, Z.~Hu, and X.~Sun, ``Towards efficient diffusion-based image editing with instant attention masks,'' in \emph{Proceedings of the AAAI Conference on Artificial Intelligence}, no.~7, 2024, pp. 7864--7872.

\bibitem{zhuang2025task}
J.~Zhuang, Y.~Zeng, W.~Liu, C.~Yuan, and K.~Chen, ``A task is worth one word: Learning with task prompts for high-quality versatile image inpainting,'' in \emph{European Conference on Computer Vision}.\hskip 1em plus 0.5em minus 0.4em\relax Springer, 2025, pp. 195--211.

\bibitem{ju2024brushnet}
X.~Ju, X.~Liu, X.~Wang, Y.~Bian, Y.~Shan, and Q.~Xu, ``Brushnet: A plug-and-play image inpainting model with decomposed dual-branch diffusion,'' \emph{arXiv preprint arXiv:2403.06976}, 2024.

\bibitem{nair2024improved}
L.~Nair, ``Improved generation of synthetic imaging data using feature-aligned diffusion,'' in \emph{Proceedings of the First International Workshop on Vision-Language Models for Biomedical Applications}, 2024, pp. 25--30.

\bibitem{brooks2023instructpix2pix}
T.~Brooks, A.~Holynski, and A.~A. Efros, ``Instructpix2pix: Learning to follow image editing instructions,'' in \emph{Proceedings of the IEEE/CVF conference on computer vision and pattern recognition}, 2023, pp. 18\,392--18\,402.

\bibitem{qi2024deadiff}
T.~Qi, S.~Fang, Y.~Wu, H.~Xie, J.~Liu, L.~Chen, Q.~He, and Y.~Zhang, ``Deadiff: An efficient stylization diffusion model with disentangled representations,'' in \emph{Proceedings of the IEEE/CVF conference on computer vision and pattern recognition}, 2024, pp. 8693--8702.

\bibitem{wei2024omniedit}
C.~Wei, Z.~Xiong, W.~Ren, X.~Du, G.~Zhang, and W.~Chen, ``Omniedit: Building image editing generalist models through specialist supervision,'' \emph{arXiv preprint arXiv:2411.07199}, 2024.

\bibitem{he2016deep}
K.~He, X.~Zhang, S.~Ren, and J.~Sun, ``Deep residual learning for image recognition,'' in \emph{Proceedings of the IEEE conference on computer vision and pattern recognition}, 2016, pp. 770--778.

\bibitem{jiang2024genai}
D.~Jiang, M.~Ku, T.~Li, Y.~Ni, S.~Sun, R.~Fan, and W.~Chen, ``Genai arena: An open evaluation platform for generative models,'' \emph{arXiv preprint arXiv:2406.04485}, 2024.

\bibitem{loshchilov2017decoupled}
I.~Loshchilov, ``Decoupled weight decay regularization,'' \emph{arXiv preprint arXiv:1711.05101}, 2017.

\bibitem{avrahami2023blended}
O.~Avrahami, O.~Fried, and D.~Lischinski, ``Blended latent diffusion,'' \emph{ACM transactions on graphics (TOG)}, vol.~42, no.~4, pp. 1--11, 2023.

\bibitem{sheikh2006statistical}
H.~R. Sheikh, M.~F. Sabir, and A.~C. Bovik, ``A statistical evaluation of recent full reference image quality assessment algorithms,'' \emph{IEEE Transactions on image processing}, vol.~15, no.~11, pp. 3440--3451, 2006.

\bibitem{zhang2018unreasonable}
R.~Zhang, P.~Isola, A.~A. Efros, E.~Shechtman, and O.~Wang, ``The unreasonable effectiveness of deep features as a perceptual metric,'' in \emph{Proceedings of the IEEE conference on computer vision and pattern recognition}, 2018, pp. 586--595.

\bibitem{wang2004image}
Z.~Wang, A.~C. Bovik, H.~R. Sheikh, and E.~P. Simoncelli, ``Image quality assessment: from error visibility to structural similarity,'' \emph{IEEE transactions on image processing}, vol.~13, no.~4, pp. 600--612, 2004.

\end{thebibliography}

\vspace{11pt}

\vfill

\end{document}